\documentclass[preprint,12pt]{elsarticle}

\usepackage{amssymb}
\usepackage{amsmath}
\usepackage[T1]{fontenc}
\usepackage{graphicx}
\usepackage{booktabs}    
\usepackage{siunitx}     
\usepackage{placeins}
\usepackage{longtable} 
\usepackage{array} 
\usepackage{float} 
\usepackage{pdflscape}

\usepackage{hyperref}    

\biboptions{numbers,sort&compress}

\begin{document}

\begin{frontmatter}
\title{From Heuristics to Data: Quantifying Site Planning Layout Indicators with Deep Learning and Multi-modal Data}
\author[aff1]{Qian Cao\corref{cor1}}
\ead{qiancao@u.nus.edu}
\ead[url]{https://orcid.org/0009-0009-2308-7800}

\author[aff2]{Jielin Chen}
\ead{jielin.chen@cares.cam.ac.uk}
\ead[url]{https://orcid.org/0000-0003-0666-8725}

\author[aff3]{Junchao Zhao}
\ead{robinzhaoyy@outlook.com}
\ead[url]{https://orcid.org/0009-0002-2386-4541}

\author[aff1]{Rudi Stouffs\corref{cor2}}
\ead{stouffs@nus.edu.sg}
\ead[url]{https://orcid.org/0000-0002-4200-5833}

\cortext[cor1]{First author}
\cortext[cor2]{Corresponding author}

\address[aff1]{organization={National University of Singapore (NUS)},
            city={Singapore},
            country={Singapore}}

\address[aff2]{organization={The Cambridge Centre for Advanced Research and Education in Singapore (CARES)},
            city={Singapore},
            country={Singapore}}

\address[aff3]{organization={China Southwest Architectural Design and Research Institute Co., Ltd. (CSWADI)},
            city={Chengdu},
            country={China}}

\biboptions{numbers,sort&compress}
            
\begin{abstract}
%


The spatial layout of urban sites directly influences land-use efficiency and spatial organizational patterns. However, traditional site planning methods primarily rely on experiential judgment and single-source data analysis, limiting their capacity to systematically quantify the complexities of multifunctional urban layouts. This study proposes a Site Planning Layout Indicator (SPLI) system, a data-driven approach that integrates empirical knowledge with heterogeneous multi-source data, transforming it into structured urban spatial knowledge. This provides a foundational framework for future multimodal spatial data systems supporting urban analytics, inference, and retrieval. The SPLI systematically quantifies urban spatial characteristics by integrating multimodal data sources, including OpenStreetMap (OSM), Points of Interest (POI), building morphology, land use data, and satellite imagery. Furthermore, it extends traditional planning metrics by introducing five critical dimensions to comprehensively characterize urban functional layouts: (1) Hierarchical Building Function Classification, refining empirical classification systems into clear hierarchical structures; (2) Spatial Organization, structured quantification based on seven typical layout patterns (e.g., symmetrical, concentric, axial-oriented layouts); (3) Functional Diversity, converting qualitative assessments into measurable indicators using Functional Ratio (FR) and Simpson Index (SI); (4) Accessibility to Essential Services, integrating facility distribution and transportation networks to create comprehensive accessibility metrics; and (5) Land Use Intensity, employing Floor Area Ratio (FAR) and Building Coverage Ratio (BCR) to assess spatial utilization efficiency consistent with conventional planning standards. Additionally, this research addresses data gaps effectively through deep learning techniques such as Relational Graph Neural Networks (RGNN) and Graph Neural Networks (GNN). Experimental results indicate that the SPLI significantly enhances urban functional classification accuracy, providing a standardized framework supporting automated, data-driven spatial analytics.

\end{abstract}

\begin{graphicalabstract}
\end{graphicalabstract}

\begin{highlights}
\item Proposes a data framework for quantifying site planning layout indicators.
\item Framework integrates multi-modal spatial data for standardized urban analytics.
\item Designed to enable future LLM-based reasoning and inference on spatial layouts.
\item Employs deep learning (GNN/RGCN) to fill data gaps and enhance indicator accuracy.
\end{highlights}

\begin{keyword}

Site Planning Layout \sep Multi-Modal Spatial Data \sep Graph Neural Network \sep Relational Graph Convolutional Network (RGCN) \sep Urban Spatial Indicators \sep Geospatial Data Modeling

\end{keyword}

\end{frontmatter}

\section{Introduction}
\label{sec1}

Urbanization has become a major engine of global economic growth, contributing to over 80\% of the world's GDP \citep{stromquist2019world}. When effectively managed, urbanization not only enhances productivity and innovation but also facilitates sustainable development. However, rapid urbanization brings substantial challenges, including surging demands for affordable housing, infrastructure, and basic services. Once established, the physical layouts and land-use patterns of cities are challenging to modify quickly, often resulting in inefficient and disorderly urban expansion. Studies indicate that urban land consumption is growing 50\% faster than population growth, with an additional 1.2 million square kilometers of built-up areas expected globally \citep{angel2011making}. With rapid advancements in big data and artificial intelligence, effectively characterizing and analyzing urban morphology and establishing an adaptive indicator framework capable of capturing complex urban systems to support dynamically evolving architectural design and planning strategies has become a critical research direction.

Traditional site planning methods primarily rely on planners' empirical judgment and regulatory requirements, employing static functional zoning strategies that lack adaptability to the dynamic evolution of urban spaces \citep{lynch1984reconsidering,batty2013new, ledewitz1991social}. This limitation restricts precise prediction and adjustment of spatial relationships within urban functional layouts, negatively impacting residents' travel patterns, economic agglomeration effects, and efficient resource allocation. Additionally, conventional planning typically undergoes quantitative evaluation only after implementation, lacking systematic mechanisms for dynamic optimization, thus constraining the adaptability and flexibility of functional layouts. Recently, advances in big data and AI technologies have significantly improved the quantitative analysis and optimization of urban functional layouts, making planning processes more dynamic, intelligent, and accurately responsive to evolving urban contexts \citep{batty2018inventing}.

To address the current issues of insufficient and incomplete high-quality urban spatial data, deep learning methods such as Graph Neural Networks (GNN), especially Relational Graph Convolutional Networks (RGCN), have shown significant potential in data imputation and urban functional layout analysis. These data-driven models effectively infer and complete missing information by mining complex spatial and attribute relationships among multimodal data, thereby enhancing the integrity and analytical accuracy of spatial data \citep{wu2020comprehensive}. Furthermore, GNN and RGCN enable deep integration of multimodal data, transforming experiential data into structured and vectorized forms, thus opening new pathways for knowledge representation and intelligent reasoning in urban planning. Nevertheless, current research still faces limitations, including insufficient maturity in multi-source data integration methods, the absence of standardized spatial data representations, and inadequate modeling of dynamic urban functional layout processes, which collectively constrain the in-depth application of data-driven approaches in urban planning \citep{li2017urbanization}.

Given this context, this study proposes a data-driven Site Planning Layout Indicator (SPLI) system, aiming to establish a structured indicator framework capable of systematically quantifying spatial characteristics across diverse land parcels and addressing data incompleteness through multimodal data integration and deep learning approaches. Based on traditional planning indicators, SPLI introduces five key dimensions to comprehensively characterize the complexity and multi-scale features of urban functional layouts. These dimensions not only enhance the structured descriptive capability of urban spatial organization patterns but also standardize and systematize the transformation of multimodal data into knowledge graphs. By integrating RGCN and other spatial analytical techniques, SPLI effectively supports building function classification and site layout pattern recognition. Experimental results demonstrate that SPLI clearly characterizes functional layout features across different land-use types and regions, providing a solid theoretical foundation and technical support for data-driven spatial planning and intelligent reasoning.

\section{literature review}
\label{sec2}

\subsection{ Site Planning and Functional Layout}
\label{subsec1}

Urban functional layout plays a pivotal role in urban planning, significantly influencing residents' travel patterns, social interactions, and overall urban accessibility and livability \citep{ledewitz1991social,lynch1984reconsidering,zhang2023inferring,zhong2014inferring}. Early urban planning theories primarily emphasized the physical organization of space and its implications for social behavior. For example, Space Syntax reveals how spatial accessibility shapes social interaction through analyzing the relationships between buildings and street networks, while Cognitive Mapping theory explores how individuals construct mental representations of space using environmental cues. These theoretical frameworks have provided essential conceptual tools for understanding urban spatial structures and have significantly informed urban planning practices. Nevertheless, despite classical urban design theories highlighting the importance of spatial organization, connectivity, and accessibility for urban functions and social interactions \citep{ledewitz1991social, bacon1976design, peng1983theory}, urban planning has predominantly relied on empirical judgments rather than systematic quantitative analyses due to the lack of robust quantitative methodologies.

Recently, with the advancement of data-driven methodologies, urban functional layout research has increasingly focused on the finer scale of plots to address complex urban functional demands. In
contrast to traditional single-use zoning approaches, the emergence of Mixed-Use Development concepts has offered new perspectives for optimizing land-use efficiency and enhancing urban vitality (Urban Complexity and Spatial Heterogeneity). Functional diversity, a key indicator for assessing the adaptability of urban spaces, has become a central topic in urban functional layout research. Unlike earlier qualitative approaches, current studies quantitatively measure functional diversity using metrics such as Functional Ratio (FR) and Simpson Index (SI), which evaluate the balance among different building functions within individual plots. These methods provide essential data support for optimizing building functional layouts and open new possibilities for intelligent spatial reasoning and large-scale urban planning decisions. Nevertheless, despite progress in evaluating urban functional complexity, existing methods encounter limitations in handling multi-scale and highly heterogeneous urban data, thus highlighting the urgent need for more sophisticated data fusion and analytical techniques.

\begin{figure}[t]
    \centering
    \includegraphics[width=1\linewidth]{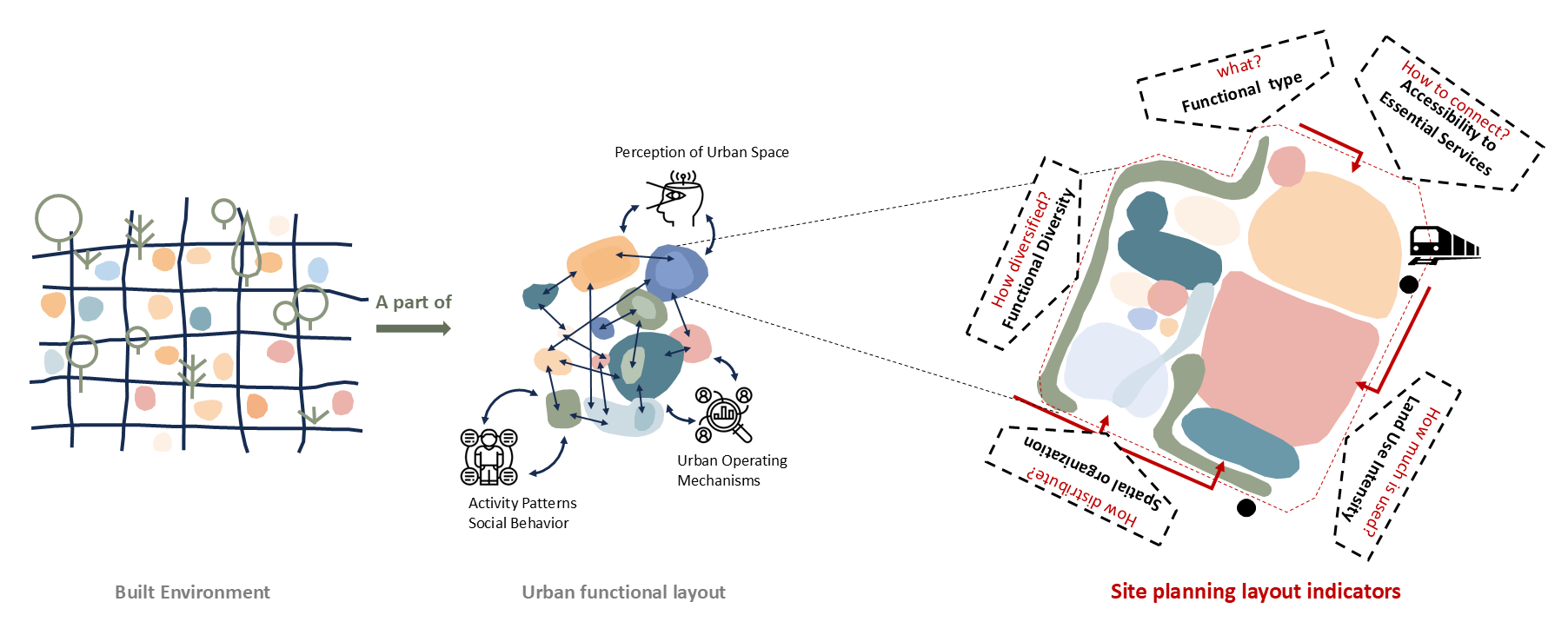}
    \caption{The scope of Site Planning Layout Indicators}
    \label{fig:The_scope_of_Site_Planning_Layout_Indicators}
\end{figure}

In site planning, Urban Morphology, as the core of physical spatial structure, determines spatial organization, land-use intensity, and accessibility within plots \citep{jhaldiyal2018urban,biljecki2022global}, as shown in Fig.~\ref{fig:The_scope_of_Site_Planning_Layout_Indicators}. Traditional studies primarily quantify urban morphology through indicators such as Building Coverage Ratio (BCR), Floor Area Ratio (FAR), and Building Height, combined with building layout classifications like Symmetrical Layout, Axis-Guided Layout, and Centripetal Layout (Architecture Space Composition Theory) to analyze spatial organizational characteristics of different urban functional areas. However, recent integration of satellite imagery, morphological data, and machine learning methods has significantly enhanced the accuracy of automated identification and classification of building forms. Additionally, applying Graph Neural Networks (GNN) has overcome limitations of conventional geometric analyses, enabling the quantification of complex spatial relationships between building clusters and providing intelligent data-driven solutions for building function classification and urban functional layout optimization \citep{atitallah2020leveraging}.

\subsection{Deep Learning Applications in Urban Planning}
\label{subsec2}

With the increasing complexity of urban systems, traditional GIS and CAD methods, although useful in describing architectural morphology and spatial layouts, encounter significant limitations in handling complex spatial relationships, dynamic interactions, and multi-scale data integration in urban environments \citep{dollner2007integrating}. Given the multidimensional nature of urban settings, encompassing social, economic, and environmental variables, conventional rule-based and statistical analysis approaches often struggle to effectively model the dynamic and intricate interactions within urban spaces \citep{albeverio2007dynamics}. Recently, deep learning techniques, particularly Graph Neural Networks (GNN), have been widely applied in urban planning, offering new data-driven methodologies for tasks such as functional classification, urban morphology analysis, urban operation optimization, and facility siting \citep{wu2020comprehensive}.

GNN applications in urban planning primarily include urban functional classification, urban morphology analysis, urban operational optimization, and facility location optimization. For urban functional classification, GNN significantly improves the accuracy of building function prediction by learning spatial relationships among buildings, roads, and facilities, enabling refined urban functional zoning based on spatial topology and Points of Interest (POI) data \citep{zhang2023inferring, zhong2014inferring}. In urban morphology analysis, GNN automatically extracts building morphological indicators, such as building height, density, and Floor Area Ratio (FAR), analyzes spatial organization patterns of building clusters, and evaluates urban development patterns and land-use efficiency \citep{jhaldiyal2018urban,biljecki2022global}. Regarding urban operation optimization, GNN models dynamic networks involving pedestrian flow, logistics, and traffic flow, providing essential data support; for instance, Spatial-Temporal Graph Convolutional Networks (STGCN) capture spatial and temporal dependencies within traffic networks, optimizing urban traffic management and enhancing transportation efficiency \citep{yu2017spatio}. For facility location optimization, GNN assesses spatial accessibility of critical infrastructures such as hospitals, shopping centers, and transportation hubs, optimizing site selection strategies within urban spatial networks to maximize service coverage and resource allocation effectiveness.

Despite GNN's strong spatial modeling capabilities in urban planning, urban environmental data typically exhibit high heterogeneity, characterized by significant structural differences among diverse spatial data types (e.g., building functions, road networks, POI, land use, and satellite imagery). To overcome this challenge, Relational Graph Convolutional Networks (RGCN) have been introduced to enhance GNN's capability in handling heterogeneous data. RGCN incorporates different types of edge weights, effectively capturing complex associations among buildings, facilities, roads, and land uses, thus improving accuracy and generalization in urban function prediction \citep{kipf2016semi}. In urban planning, RGCN has been effectively utilized for multimodal data fusion, integrating multiple data sources such as OpenStreetMap (OSM), POI distribution, building morphologies, and satellite imagery, enabling precise modeling of urban functional layouts and supporting data-driven spatial planning decisions \citep{batty2013big}. Compared to conventional GNN methods, RGCN provides fine-grained modeling of various spatial entities and their relationships, enhancing the integration of heterogeneous data and offering new directions for intelligent optimization in urban planning.

\subsection{Research Gaps}
\label{subsec3}

The spatial layout of urban sites directly influences land use efficiency and spatial organization. However, traditional site planning methods primarily rely on planners’ empirical judgment and single-source data analysis, making it challenging to systematically quantify the complexity of multifunctional urban layouts \citep{ratti2004space, porta2006network}. While data-driven approaches have advanced functional classification, spatial layout analysis, and facility siting optimization \citep{batty2013big}, existing research still faces challenges in multimodal data integration, spatial knowledge representation, and functional layout inference.

First, urban planning research lacks a standardized framework that systematically integrates multimodal data sources such as OpenStreetMap (OSM), Points of Interest (POI), building morphology, land use, and satellite imagery while utilizing knowledge graphs for structured indicator systems. Current methods often rely on single-source data or static analyses, failing to capture the dynamic and multi-scale characteristics of urban functional layouts, which limits classification accuracy \citep{li2019spatial}. Moreover, inconsistencies in temporal scales, spatial resolutions, and semantic standards across data sources hinder uniform representation, complicating cross-regional and multi-scale urban spatial analysis \citep{su2021urban}.

Second, despite the growing adoption of knowledge graphs in data integration, urban planning still lacks a standardized knowledge framework for functional layouts and spatial organization, leading to poor data interoperability across studies \citep{jeong2020city, viola2019mapping}. Traditional GIS methods primarily use vector-based static classification, relying on land use labels or POI categorization to represent building functions but fail to capture complex functional relationships, such as clustering effects, competitive dynamics, or functional complementarity among buildings \citep{dollner2007integrating}. This limitation results in fragmented urban function layout analysis, restricting its potential for intelligent urban planning applications.

Furthermore, existing research on urban spatial data completion remains inadequate. Incomplete or outdated urban function data negatively impacts analytical accuracy. Recent advancements in deep learning, particularly Graph Neural Networks (GNN), have demonstrated significant potential in urban function prediction and spatial analysis \citep{wu2020comprehensive}. Relational Graph Convolutional Networks (RGCN) have been applied to functional classification and layout pattern recognition due to their advantages in heterogeneous data fusion \citep{kipf2016semi}. By incorporating different edge weights, RGCN captures intricate interactions among buildings, enhancing classification accuracy. However, current studies mainly focus on individual building function classification, lacking systematic modeling of urban spatial organization at the plot scale. This gap limits the ability to capture hierarchical relationships and the dynamic evolution of urban functions.

Finally, although deep learning and GNN have improved building function classification, their potential in geospatial reasoning remains underutilized. Most GNN models are limited to static classification tasks and do not infer functional interactions between buildings, such as the interdependencies among office, commercial, and residential functions. Simultaneously, Large Language Models (LLMs) have recently exhibited strong generalization capabilities, but their reasoning capacity in urban planning remains constrained due to a lack of structured spatial knowledge. The Retrieval-Augmented Generation (RAG) paradigm has been proposed to provide structured knowledge frameworks for LLMs, improving their inference capabilities in building function prediction, facility siting, and spatial optimization \citep{lewis2020retrieval, guu2020retrieval, yu2025spatial}. However, no systematic research has yet explored how a standardized spatial knowledge system based on quantitative urban data can enhance LLMs’ reasoning performance in urban function analysis.

\subsection{Research Motivation and Study Overview}
\label{subsec4}

This study addresses three key challenges in current urban function classification. First, existing methods lack unified quantitative standards, resulting in inconsistencies in representation and limited comparability across studies. Second, urban function data are often incomplete or outdated, affecting analytical accuracy. Third, current approaches lack standardized storage and retrieval mechanisms, which limits data scalability and applicability.

Section 3 introduces a scalable indicator system that integrates OSM, POI, building morphology, land use, and satellite imagery to systematically represent urban functional layouts and spatial characteristics. In Section 4, we explore how Graph Neural Networks (GNN), Relational Graph Convolutional Networks (RGCN), and heterogeneous data fusion techniques can improve the completeness and reliability of urban function classification. Section 5 demonstrates the proposed framework using empirical data and analysis. Finally, Section 6 reflects on the study’s limitations and outlines future directions, particularly the development of an indicator system that incorporates knowledge graphs and retrieval-augmented generation (RAG) to enhance LLM inference in urban function classification, spatial layout optimization, and facility siting. The paper concludes in Section 7.

\section{SPLI Framework and Indicator System}
\label{sec3}

\subsection{Rationale for Plot-Level Indicator System}
\label{subsec1}

Site planning is the core process of optimizing urban functional layouts, with its key challenge lying in accurately characterizing spatial features at the plot level to support rational functional organization, spatial layout, and resource allocation \citep{shi2023defining}. However, existing studies primarily focus on macro-scale urban functional zoning, lacking a detailed depiction of functional organization within individual plots, which limits decision-making support in functional layout optimization and planning implementation. This study proposes the Site Planning Layout Indicator (SPLI), which conducts functional classification and spatial feature quantification at the plot level, aiming to establish a standardized quantitative method that supports data-driven urban planning decisions.

The plot level is chosen as the fundamental unit of the SPLI system primarily due to its key role in urban planning and architectural design. As the basic unit of urban spatial organization, plots carry building morphology, land use, and functional layouts, making them an ideal scale to bridge urban planning and architectural design \cite{batty2013new}. Compared to analyses at the street block or district level, plot-scale studies can more accurately capture spatial land-use characteristics. Especially in the context where mixed-use development has become the mainstream mode of urban spatial optimization, plot-level analysis can better reveal the internal functional configuration of multi-functional land use.

Additionally, the plot scale allows for strong data integration and is well-suited for multi-modal data fusion. Urban spatial research involves diverse datasets, including GIS data, points of interest (POI), building morphology, land use plans, and satellite remote sensing images. Plot-level analysis enables the fusion of these multi-source datasets at a fine-grained scale and, through Relational Graph Convolutional Networks (RGCN), learns the latent patterns between building functions and spatial layouts, thereby enhancing analysis accuracy and applicability \citep{wu2020comprehensive}.

Furthermore, plot-level analysis enhances the operational feasibility of urban planning decisions. Urban planning often requires functional adjustments at the plot scale, such as the spatial allocation of residential, commercial, educational, and healthcare functions, as well as the optimization of key planning parameters, including building height, density, and accessibility. By quantifying functional layouts through the SPLI system, urban planners can obtain more instructive data support to ensure the rationality and sustainability of urban functional layouts. Thus, the SPLI system, based on plot-level quantitative analysis, not only fills the gap in fine-grained characterization of multi-functional land use in existing research but also provides a data foundation for future intelligent urban planning, promoting the deeper application of data-driven methods in architectural design and urban planning.

\subsection{Data Sources}
\label{subsec2}

This study constructs the Site Planning Layout Indicator (SPLI) based on multimodal data integration to quantitatively evaluate the complexity of urban functional layouts. In terms of data selection, the SPLI system follows these criteria: (1) ensuring multidimensionality to cover key planning dimensions such as building function, spatial organization, facility accessibility, and land use intensity; (2) emphasizing timeliness and spatial consistency to ensure alignment between different data sources and accurately represent the current conditions of the study area; (3) ensuring data accessibility by selecting public and authoritative urban planning datasets to enhance the generalizability of the methodology.

On this basis, the SPLI system integrates global site planning standards, geographic information data, building morphology data, land use data, and satellite imagery. Regarding global site planning standards, this study reviews relevant regulations from multiple countries and regions, including China’s Urban and Rural Planning Law and Urban Residential Area Planning and Design Standard (GB 50180-2018), the United States’ International Building Code (IBC), Zoning Ordinances, and Unified Development Ordinances (UDOs), as well as Singapore’s Urban Redevelopment Authority Master Plan and Building Control Regulations. These regulations define key indicators such as floor area ratio (FAR), building coverage ratio (BCR), building density, height restrictions, setback distances, and green plot ratio (GnPR), providing the theoretical basis for the SPLI system.

\begin{figure}[t]
    \centering
    \includegraphics[width=1\linewidth]{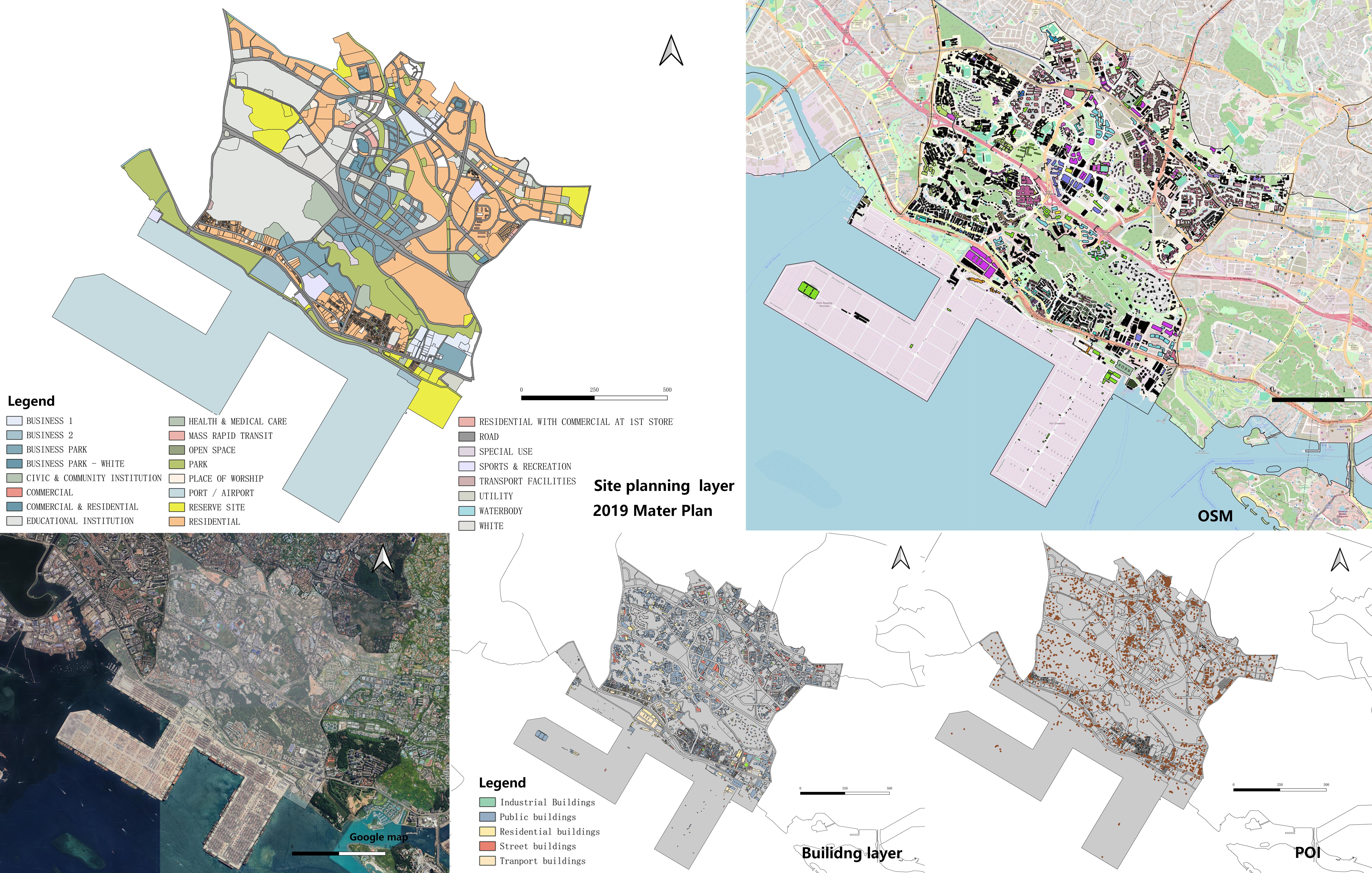}
    \caption{Study area and data types}
    \label{fig:Studyareaanddatatype}
\end{figure}

For spatial data, this study utilizes OpenStreetMap (OSM) to obtain street networks, plot boundaries, and building information and employs points of interest (POI) data to identify urban facility types and distribution patterns, as shown in Fig.~\ref{fig:Studyareaanddatatype}. Building morphology data, including building height, footprint area, and geometric morphology, are primarily used to analyze spatial organization and classify building functions. Land use data are sourced from Singapore’s 2019 Master Plan to ensure the classification system aligns with official planning standards. In addition, satellite imagery is incorporated to extract building density, geometric morphology, and environmental characteristics, enhancing the accuracy of morphological analysis. Through multimodal data integration, the SPLI system systematically quantifies site planning layout characteristics, providing data support for intelligent urban functional analysis.

\subsection{Study Area}
\label{subsec3}

This study selects Queenstown, as shown in Fig.~\ref{fig:Study area and data type}, Singapore, as the research area due to its representativeness in terms of urban renewal, mixed-use land distribution, and data availability. As Singapore’s first satellite town, Queenstown has undergone multiple rounds of urban renewal since its development in the 1950s. Its diverse architectural typologies and functional layouts, encompassing mature public housing (HDB), modern condominiums, and commercial complexes, provide a comprehensive testing ground for evaluating the applicability of the SPLI system in complex urban environments. Additionally, the integration of residential, commercial, industrial, educational, healthcare, and public spaces within this area makes it an ideal case study for multi-functional urban land use research.

Queenstown offers high data availability, as the Singaporean government provides high-precision GIS datasets, including building morphology, land use, POI, and satellite imagery, forming a robust data foundation for the development and validation of the SPLI system.  The spatial structure of the study area can be categorized into five primary typologies: high-density residential zones comprising a mix of HDB estates, private housing, and condominiums; commercial and office areas composed of shopping malls, office buildings, and mixed-use developments; educational and research zones clustered around institutions such as Singapore Polytechnic and the National University of Singapore; green and open spaces including waterfront parks, community plazas, and recreational grounds; and port and logistics zones, located in the western part of the study area, characterized by concentrated warehouse and logistics facilities that reflect typical transport and industrial support functions. The complex functional composition of Queenstown provides a rich testing ground for evaluating the applicability and robustness of the SPLI framework.

\subsection{SPLI Indicator System}
\label{subsec4}

The SPLI system integrates traditional urban planning indicators with data-driven analytical methods, forming a systematic quantitative framework. Building upon conventional data indicator systems, SPLI introduces five core dimensions: functional typology, spatial organization, functional diversity, accessibility, and land use intensity, to characterize the spatial attributes of urban functional layouts, as shown in Table \ref{tab:SPLI_Indicators}.

Functional Typology adopts a three-tiered classification method, including building category, functional type, and specific building function. This classification framework ensures both logical clarity and alignment with Singapore’s Master Plan and relevant regulations. By employing this hierarchical classification, SPLI refines the spatial distribution of different building functions, providing essential data support for building function prediction.

Spatial Organization focuses on the layout patterns of buildings, identifying seven typical configurations: absolute or approximate symmetrical layout, centripetal layout, axis-guided layout, uniform form, mixed layout, and flexible layout. These patterns are classified based on the spatial distribution of building clusters and their relationship to site boundaries, with morphological characteristics computed using building form data.

Functional Diversity is assessed by calculating the Functional Ratio (FR) and Simpson Index (SI) to measure the balance of functional distribution. The Functional Ratio quantifies the proportion of different functions within a plot, while the Simpson Index evaluates functional diversity, reflecting the balance and complementarity among different functions.

Accessibility encompasses access to daily necessities and public transit accessibility. Access to daily necessities is measured based on POI data, calculating the spatial distance between plots and key facilities such as retail stores, schools, and hospitals. Public transit accessibility is evaluated through GIS network analysis, assessing walking accessibility to bus stops and metro stations, thereby quantifying the convenience of urban functional layouts.

Land Use Intensity is quantified using Floor Area Ratio (FAR) and Building Coverage Ratio (BCR) to measure development density at the plot level. FAR is calculated as the ratio of total floor area to plot area, indicating the intensity of land development, while BCR represents the proportion of building footprint to plot area, assessing the efficiency of spatial utilization. Widely applied in urban planning and land management, these indicators enable the SPLI system to effectively characterize urban spatial development patterns.

\renewcommand{\arraystretch}{0.5} 
\scriptsize
\begin{longtable}{
    >{\centering\arraybackslash}p{3.5cm}
    >{\centering\arraybackslash}p{3cm}
    >{\centering\arraybackslash}p{2cm}
    >{\centering\arraybackslash}p{3.5cm}
}
\caption{Site Planning Layout Indicators} \label{tab:SPLI_Indicators} \\
\toprule
\textbf{Indicator} & \textbf{Example Values} & \textbf{Data Type} & \textbf{Description} \\ 
\midrule
\endfirsthead 
\toprule
\textbf{Indicator} & \textbf{Example Value} & \textbf{Data Type} & \textbf{Description} \\ 
\midrule
\endhead

\multicolumn{4}{l}{\textbf{Building Level Indicators}} \\ 
\midrule
\textbf{General Indicators} & & & \\
Building Name & B1620, B1849 & String & Unique ID \\
Latitude & 1.277, 1.286 & Decimal & Latitude coordinate \\
Longitude & 103.799, 103.781 & Decimal & Longitude coordinate \\
Foot Area & 366.87, 437.78 & Float & Footprint area (sqm) \\
Perimeter & 79.432, 113.345 & Float & Perimeter length (m) \\
Height & 3, 27 & Integer & Building height (m) \\
Number of Floors & 1, 5 & Integer & Total floors \\
Orientation & 203.39, 139.2 & Float & Angle from true north (°) \\ 
\midrule
\textbf{Function-related Indicators} & & & \\
Building Category & Public, Street & Categorical & Broad classification \\
Building Type & Office, Residential & Categorical & Based on planning regulations \\
Building Function & Corridor, Hospital & Categorical & Usage classification \\
Building Form & Point, Slab & Categorical & Geometric type \\ 
\midrule
\multicolumn{4}{l}{\textbf{Site Planning Level Indicators}} \\ 
\midrule
\textbf{Site Layout Indicators} & & & \\
MP Name & kml\_8644, kml\_109008 & String & Plot ID in the master plan \\
Land Use & BUSINESS 1, UTILITY & Categorical & Land use type \\
Plot Area & 47712.31, 118330.84 & Float & Plot size (sqm) \\
Subzone & PASIR PANJANG 2 & String & Subzone classification \\ 
\midrule
\textbf{Spatial Organization} & & & \\
Layout Pattern & Uniform, Centripetal & Categorical & Internal spatial arrangement \\ 
\midrule
\textbf{Functional Diversity} & & & \\
Functional Ratio (FR) & 0.022, 0.9116 & Float & Function proportion in a site \\
Simpson Index (SI) & 0.5853, 0.1612 & Float & Function diversity score \\ 
\midrule
\textbf{Accessibility to Services} & & & \\
Public Transit Access (PTA) & 0, 0.05 & Float & Access to public transport \\
Connectivity Index (CI) & -- & Float & Site connectivity measure \\ 
\midrule
\textbf{Land Use Intensity} & & & \\
FAR & 2.57, 4.53 & Float & Floor area ratio \\
BCR & 0.35, 0.19 & Float & Ground coverage ratio \\ 
\bottomrule
\end{longtable}

\section{Implementation}
\label{sec4}

\subsection{Data Preprocessing}
\label{subsec1}

Before constructing the SPLI system, multimodal data must be standardized to ensure consistency and usability. This study adopts Singapore’s projected coordinate system (EPSG:3414) for data format conversion and coordinate alignment, enabling the integration of data from various sources within a unified spatial framework.

First, the street network, plot boundaries, and building data from OpenStreetMap (OSM) are cleaned and formatted to remove redundant information. A spatial indexing method is applied to optimize data retrieval efficiency. Next, POI data undergoes standardization, extracting facility types, spatial coordinates, and relevant attributes, while a geographically weighted adjustment method is employed to address uneven POI distribution. Additionally, building morphology data, including building height, footprint area, and geometric features, are processed for missing values and standardized to ensure computational accuracy.

In data preprocessing, addressing missing data is also crucial. For plots where building function information is unavailable, this study infers missing functions by combining surrounding POI distributions and land use data. Furthermore, for building density, green coverage ratio, and other features extracted from satellite imagery, image classification techniques are applied to enhance accuracy, minimizing the impact of data noise on subsequent computations.

\subsection{Hierarchical Building Function Classification}
\label{subsec2}

Building function classification is a fundamental aspect of site planning and urban design, directly influencing spatial organization, facility allocation, and land use efficiency. However, existing publicly available data on building function classification have limitations that hinder their applicability for detailed urban studies. To address this issue, this study systematically organizes building function classification based on Singapore’s 2019 Master Plan and relevant regulations, including the Building Control Regulations, Code of Practice on Buildable Design, and Development Control Plan. A hierarchical building function indicator system is developed to enhance the accuracy and applicability of building function analysis, as shown in Table~\ref{tab:building_summary1}; for the full classification structure, see AppendixA (Table\ref{tab:building_summary}).

This study employs a three-tiered building function classification system to ensure logical clarity and compliance with Singapore’s urban planning standards. The third tier (Building Function) is categorized based on the land-use types defined in the Singapore Master Plan, specifying permissible building functions such as Student Hostel and Shophouse. Further classification is conducted based on specific usage attributes, identifying ancillary structures such as Corridor and Annex to refine building functions and enhance classification accuracy. At the second tier, functions from the third tier are merged based on big data computational needs and user demographics, forming broader functional groups such as Office \& Industrial Buildings and Residential Buildings. This classification reflects variations in building types based on different functional demands. The first tier (Building Category) is derived from the second tier, following macro-level functional attributes and architectural theory-based classification, encompassing Public Buildings, Residential Buildings, Street Buildings, and Industrial/Transport Buildings, covering major urban architectural types. This classification system ensures alignment with Singapore’s statutory planning framework, while incorporating architectural classification methodologies to maintain logical rigor and operational feasibility. The multi-tiered structure is designed to support spatial analysis and optimization of building functions, catering to different levels of built environment analysis.

\begin{figure}[t]
    \centering
    \includegraphics[width=0.75\linewidth]{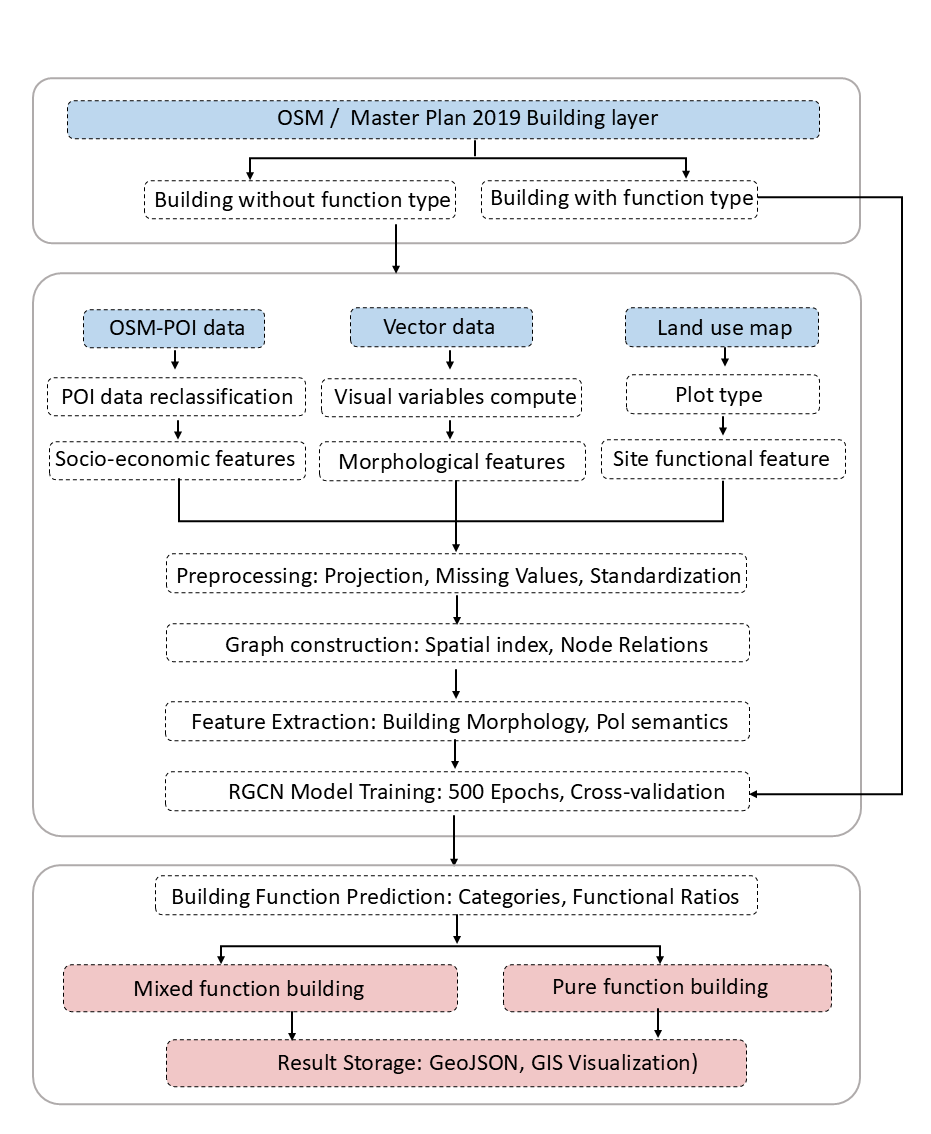}
    \caption{Building Function classification workflow}
    \label{fig:BuildingFunctionClassificationWorkflow}
\end{figure}

In building function prediction, this study integrates land use data, building morphology data, and points of interest (POI) data, employing a Relational Graph Convolutional Network (RGCN) to classify building functions at the second hierarchical level, as shown in Fig. \ref{fig:BuildingFunctionClassificationWorkflow}. First, GeoPandas is utilized to process land use data, aligning and standardizing it with building morphology and POI data to ensure spatial consistency. Building morphology data include key attributes such as building height, footprint area, and volumetric characteristics, with missing values imputed to enhance data completeness. POI data are used to extract facility types, ratings, and geographic coordinates to enrich the spatial representation of building functions.

During model construction, a spatial indexing method (R-tree) is employed to establish adjacency relationships between buildings, with an adaptive spatial threshold to ensure the validity of functional associations. A Graph Neural Network (GNN) framework is then applied, where RGCN is used for relational modeling, simultaneously learning building spatial distribution patterns and reinforcing contextual dependencies among building functions. Unlike traditional classification methods that rely on geometric or statistical rules, RGCN adopts an end-to-end learning approach, adaptively extracting high-dimensional spatial features and capturing cross-parcel functional interactions, thereby enhancing classification accuracy.

To further refine feature representation, this study employs normalized multimodal data, ensuring the integration of building morphology and POI data across different scales to improve model stability and generalizability. The training process incorporates a multi-layer RGCN architecture, combining ReLU activation functions and Dropout regularization to prevent overfitting while maintaining model expressiveness. The model is trained for 500 epochs, with parameter optimization and cross-validation to assess classification accuracy and ensure robustness and applicability.

Finally, the predicted results are stored in a GeoJSON file and integrated with GIS data, enabling the visualization of building function classification and facilitating further analysis of spatial distribution patterns for urban functional layout optimization. Based on the second hierarchical classification, building morphology characteristics, and POI data, additional refinements define corridors (length-to-width ratio greater than 3, width less than 5, and connecting two buildings), different types of religious buildings, and student accommodations within educational facilities, forming the third hierarchical level. Furthermore, considering architectural design strategies, the first hierarchical level is derived from the second level, establishing a more structured and systematic building function hierarchy.

\renewcommand{\arraystretch}{0.5}
\scriptsize
\begin{longtable}{
    >{\centering\arraybackslash}p{4.2cm}
    >{\centering\arraybackslash}p{2.8cm}
    >{\centering\arraybackslash}p{1.0cm}
    >{\centering\arraybackslash}p{2.1cm}
    >{\centering\arraybackslash}p{1.5cm}
}

\caption{Summary of the three-tier building function classification (partial list; see Appendix Table~\ref{tab:building_summary} for full version)}
\label{tab:building_summary1}
\\
\toprule
\textbf{Level 3 Building function} & \textbf{Level 2 Building type} & \textbf{Level 2 code} & \textbf{Level 1 Buildings category} & \textbf{Land use} \\
\midrule
\endfirsthead
\toprule
\textbf{Level 3 Building function} & \textbf{Level 2 Building type} & \textbf{Level 2 number} & \textbf{Level 1 Buildings category} & \textbf{Land use} \\
\midrule
\endhead
Residential-Housing Units-HDB Properties-annex-corridor & Residential-Housing Units-HDB Properties-annex & B3' & Residential buildings & Residential \\
\cmidrule(lr){1-1} 
Annex &  &  &  &  \\
\cmidrule(lr){1-4} 
Flats & Residential-Housing Units-HDB Properties & B3 &  &  \\
\cmidrule(lr){1-3} 
Annex & Residential-Housing Units-Condominiums and Other Apartments-annex & B2' &  &  \\
\cmidrule(lr){1-1} 
Residential-Housing Units-Condominiums and Other Apartments-annex-corridor &  &  &  &  \\
\cmidrule(lr){1-4} 
Condominium & Residential-Housing Units-Condominiums and Other Apartments & B2 &  &  \\
\midrule 
Townhouse & Residential-Housing Units-Landed Properties & B1 & Street buildings &  \\
\cmidrule(lr){1-1} 
Terrace House &  &  &  &  \\
\cmidrule(lr){1-1} 
Semi-Detached House &  &  &  &  \\
\cmidrule(lr){1-1} 
Detached House &  &  &  &  \\
\cmidrule(lr){1-1} 
Strata-Landed Housing &  &  &  &  \\
\cmidrule(lr){1-4} 
Retirement Housing & Residential-Housing Units-Condominiums and Other Apartments & B2 & Residential buildings &  \\
\cmidrule(lr){1-1} 
Serviced Apartments &  &  &  &  \\
\cmidrule(lr){1-1} 
Student Hostel &  &  &  &  \\
\midrule 
Flats with commercial uses at 1st storey & HDB with commercial uses at 1st storey & AB4 & Residential buildings & Residential with Commercial at 1st storey \\
\cmidrule(lr){1-4} 
Shophouse & Shophouse & B5 & Street buildings &  \\
\cmidrule(lr){1-4} 
Residential Developments(e.g. Flats) & Condominiums and Other Apartments with commercial uses at 1st storey & AB4' & Residential buildings &  \\
\midrule 
Mixed Commercial \&Residential development(e.g. Shopping/Hotel/Office \& Residential) & Commercial \& Residential & AB & Public buildings & Commercial \& Residential \\
\cmidrule(lr){2-3} 
 & Commercial, Office \& Residential & AAB &  &  \\
 \cmidrule(lr){2-3} 
 & Office, Residential \& Educational & ABD &  &  \\
 \cmidrule(lr){2-3} 
 & Commercial, Office , Transport facility \& Residential & AAAB &  &  \\
\midrule 

\bottomrule
\end{longtable}

\subsection{Spatial Organization Computation Method}
\label{subsec3}

The spatial organization characteristics of a site reflect the internal arrangement of buildings, which play a crucial role in site planning and architectural design. This study systematically analyzes the spatial organization characteristics of urban plots from two perspectives: building geometric classification and layout pattern recognition, aiming to elucidate the spatial configuration logic of different building clusters.

\begin{figure}[t]
    \centering
    \includegraphics[width=1\linewidth]{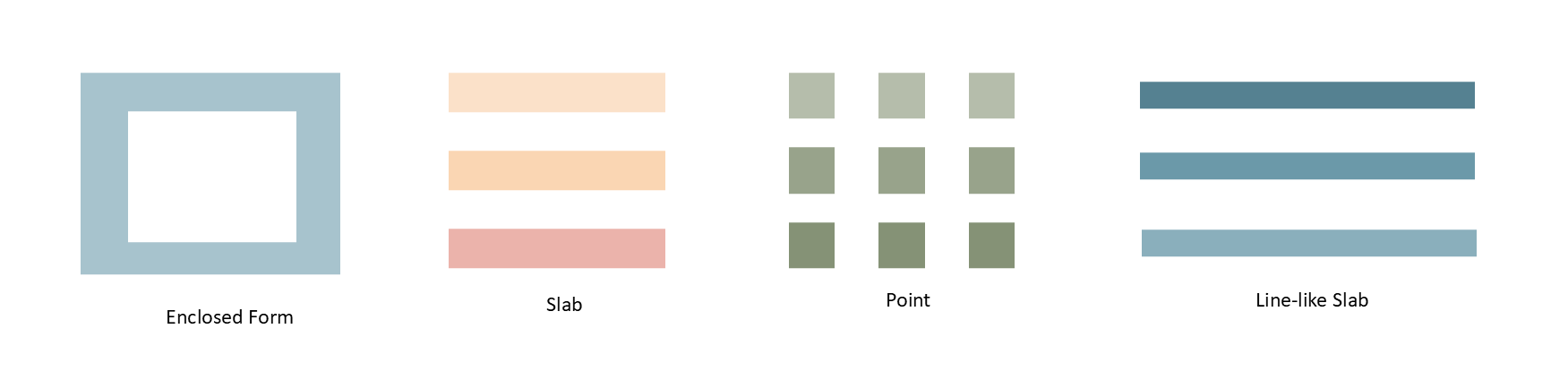}
    \caption{Illustration of the four building form categories}
    \label{fig:Buildingformillustration}
\end{figure}

\begin{figure}[t]
    \centering
    \includegraphics[width=1\linewidth]{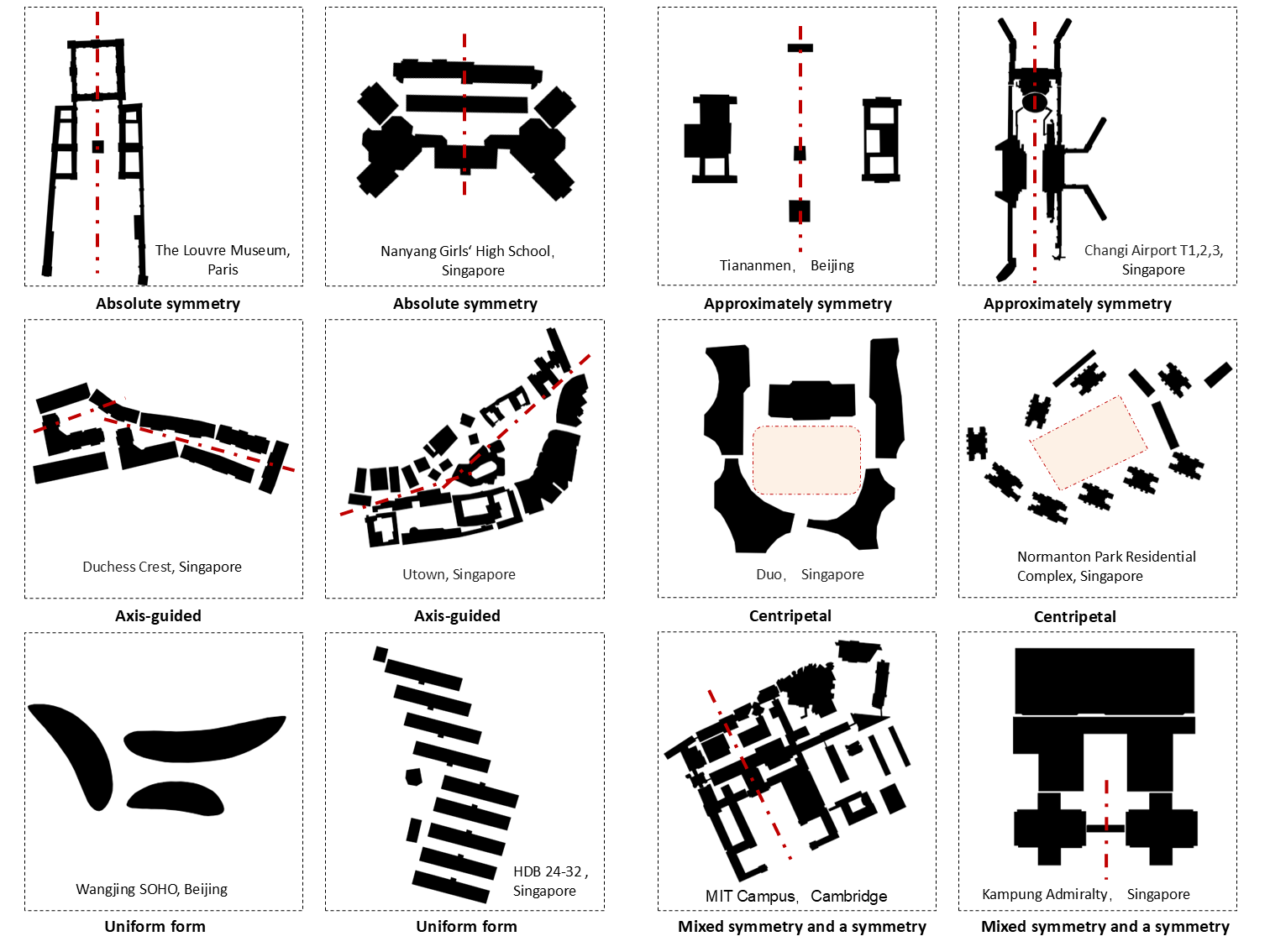}
    \caption{Illustration of six of the seven layout categories (flexible layout is omitted as it is the default category)}
    \label{fig:Theillustrationoflayoutpattern}
\end{figure}

In terms of building geometric classification, this study employs a building contour feature analysis method to categorize building forms into four types: Point, Slab, 
 Enclosed Form, and Line-like Slab  as shown in Fig.~\ref{fig:Buildingformillustration}. The classification criteria include the enclosure of the building footprint, the aspect ratio of the minimum bounding rectangle, and the presence of internal openings. For MultiPolygon buildings, the geometric attributes of each sub-polygon are computed separately, and Principal Component Analysis (PCA) is applied to derive the dominant building form category. Regarding layout pattern recognition, the study integrates building centroid calculations and symmetry analysis to classify seven typical spatial organization patterns: Absolute Symmetry, Approximate Symmetry, Centripetal Layout, Axis-Guided Layout, Uniform Form, Mixed Layout, and Flexible Layout (default, if non of the other layouts apply), as shown in Fig. \ref{fig:Theillustrationoflayoutpattern}.

\begin{figure}[t]
    \centering
    \includegraphics[width=1\linewidth]{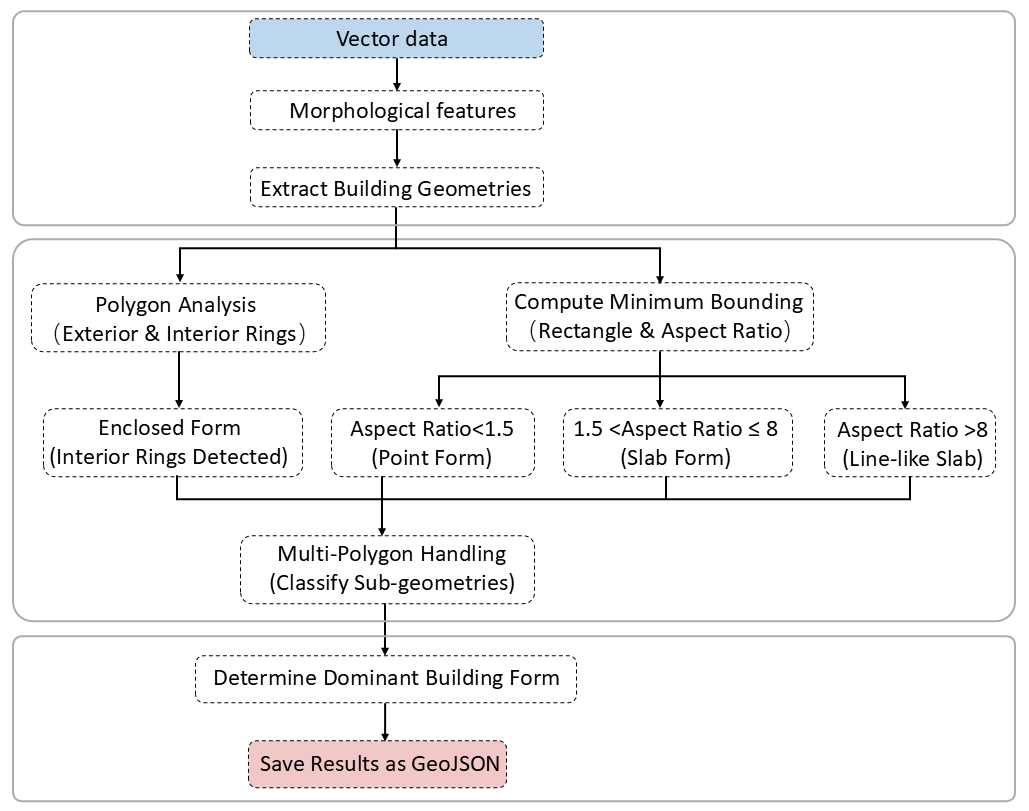}
    \caption{Building form classification workflow}
    \label{fig:Buildingformclassificationworkflow}
\end{figure}

This study employs a geometry-based analytical approach to classify building forms, aiming to standardize the representation of architectural spatial morphology. The classification process is based on the Minimum Bounding Rectangle (MBR) calculation, which evaluates the aspect ratio of each building and integrates contour characteristics for further categorization, as shown in Fig. \ref{fig:Buildingformclassificationworkflow}. Specifically, buildings containing interior rings (interiors), indicative of courtyards or enclosed spatial configurations, are classified as Enclosed Form. For buildings without interior rings, classification is determined by the aspect ratio of the MBR: those with an aspect ratio $\leq 1.5$ are categorized as Point, those with $1.5 <$ aspect ratio $\leq 8.0$ as Slab, and those with an aspect ratio $> 8.0$ as Line-like Slab. For MultiPolygon buildings, the geometric characteristics of each sub-polygon are analyzed separately, and the most frequently occurring category is assigned as the dominant building form to ensure classification consistency. Finally, a new attribute field, Form\_Type, is added to the dataset to store classification results, and the updated dataset is exported in GeoJSON format to facilitate subsequent spatial analysis and visualization.

\begin{figure}[t]
    \centering
    \includegraphics[width=0.75\linewidth]{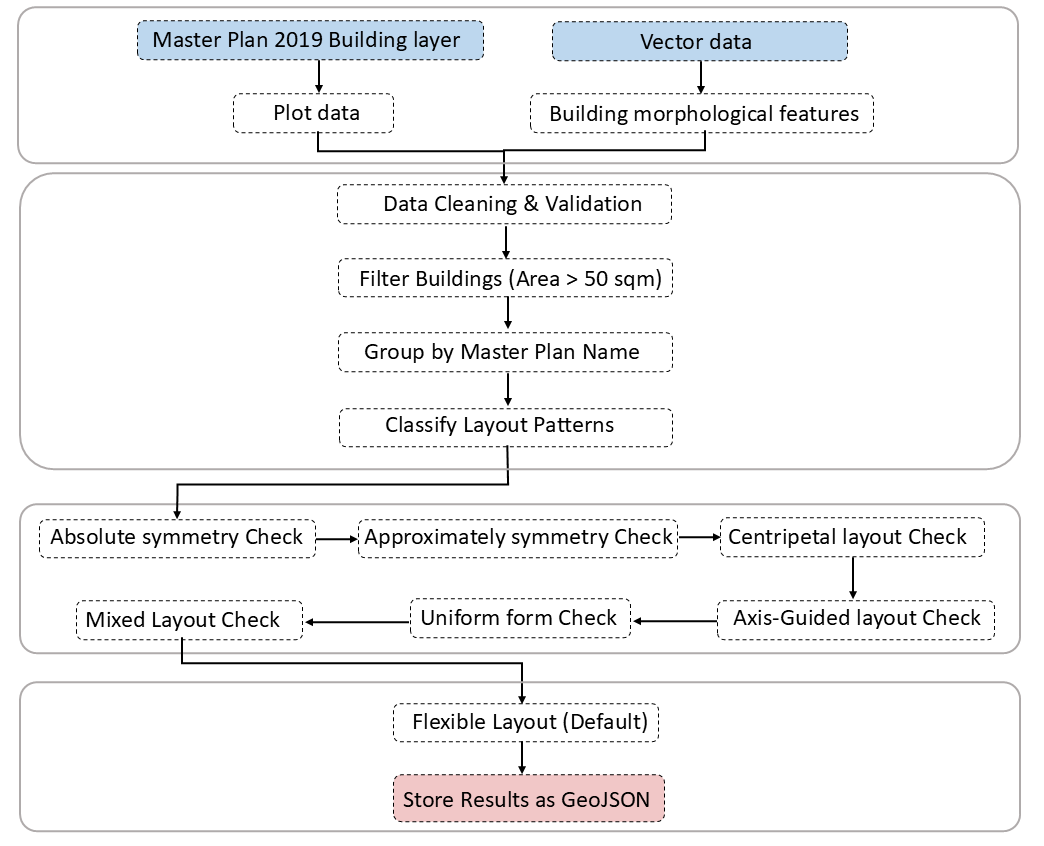}
    \caption{Spatial organization classification workflow}
    \label{fig:Spatialorganizationclassificationworkflow}
\end{figure}

The implementation of the layout classification process is based on an analysis of building morphological features, utilizing spatial computation and rule-based classification methods to systematically identify different types of building layout patterns and standardize their representation, as shown in Fig.~\ref{fig:Spatialorganizationclassificationworkflow}. The workflow consists of data acquisition and preprocessing, building filtering, geometric feature extraction, and rule-based classification to establish a robust classification system for spatial organization. First, building data and Master Plan zoning data are imported, ensuring the completeness of key attributes such as Landuse\_Name, Footprint\_Area, and Building\_Type, while verifying the validity of geometric data. To enhance classification accuracy, buildings with an area of less than 50 square meters are excluded to eliminate the influence of small auxiliary structures. Additionally, transportation-related buildings (transport facility, public transportation facility) are removed to ensure that the classification focuses primarily on residential, commercial, and office buildings.

Following data preprocessing, the study applies geometric feature extraction techniques to classify the spatial organization of buildings. The fundamental attributes computed include building centroid (Centroid), building footprint area (Area), and aspect ratio (Aspect Ratio), which serve as the basis for classification. The layout classification adopts a hierarchical decision logic, prioritizing symmetry-based classification and proceeding sequentially through the following categories: Absolute Symmetry, identified based on the numerical balance of buildings, footprint area symmetry, and centroid alignment; Approximate Symmetry, allowing for greater variance in area ratio and centroid displacement; Centripetal Layout, determined by the aspect ratio of the convex hull’s bounding rectangle, indicating an enclosed spatial configuration; Axis-Guided Layout, where buildings exhibit linear alignment along a dominant axis; Uniform Form, characterized by buildings with highly similar geometric proportions, maintaining an aspect ratio deviation within 10\%; Mixed (Symmetrical and Asymmetrical) Layout, applicable to clusters that exhibit localized symmetry but an overall irregular composition. If none of these predefined classifications apply, the layout is categorized as Flexible Layout, encompassing non-typical or dynamic spatial configurations.

Upon completion of the classification process, a new field, Layout\_Pattern, is added to the building dataset to record the assigned layout classification. For single-building plots, Layout\_Pattern is assigned a null value to ensure that classification applies only to multi-building clusters. The final classification results are stored and exported in GeoJSON format, enabling further spatial visualization and urban planning applications. By leveraging a data-driven methodology, this classification approach quantifies the spatial organization of building clusters and provides a standardized analytical framework for architectural planning and urban design.

\subsection{Functional Diversity}
\label{subsec4}

Functional diversity is a critical metric in site planning, reflecting the distribution and balance of different building functions within a given plot. This study employs the Functional Ratio (FR) and Simpson Index (SI) as quantitative indicators to systematically assess the composition and heterogeneity of land use.

The Functional Ratio (FR) quantifies the proportion of each building function relative to the total site area. It is computed by calculating the footprint area of each functional category within the site and determining its ratio to the total built-up area. This metric provides insights into the spatial dominance or secondary role of various building functions within a plot, offering a quantitative foundation for evaluating functional layout rationality.

To further assess the distributional balance of functions within a site, this study incorporates the Simpson Index (SI), calculated as follows:

\[
SI = 1 - \sum_{i=1}^{n} (r_i^2)
\]

where \( r_i \) represents the functional ratio of each building type. The SI value ranges between 0 and 1, where higher values indicate a more evenly distributed mix of functions, and lower values signify the dominance of a particular function, reflecting reduced functional diversity. This metric effectively captures the degree of functional integration within a site, providing a reference for mixed-use development and urban vitality assessments.

By computing these two indicators, this study quantitatively characterizes the functional composition of different plots and evaluates their spatial rationality, supporting the optimization of building layouts and urban planning strategies. The computed results are stored in GeoJSON format, facilitating further analysis and visualization within GIS platforms.

\subsection{Accessibility to Essential Services}
\label{subsec5}

\subsubsection{Access to Daily Necessities}

This study employs a facility-based accessibility computation method to evaluate the connectivity index of each plot. The approach consists of several key steps. First, the input GeoJSON file is loaded and preprocessed to ensure geometric data integrity, with building function names standardized to match predefined facility categories such as hospitals, schools, supermarkets, parks, and convenience stores. The centroid of each plot is then calculated and projected onto a metric coordinate system to facilitate distance computation.

Subsequently, all facilities adjacent to each plot are identified within a predefined search radius (5 kilometers). The distance between each facility and the corresponding plot centroid is calculated, and a weighting system is applied based on facility importance. Specifically, hospitals are assigned the highest weight (5), followed by schools (4), supermarkets (3), parks (2), and convenience stores with the lowest weight (1). The contribution of each facility is determined by the inverse relationship between its assigned weight and its distance from the plot, ensuring that both proximity and facility significance influence the computed connectivity index. 

The final connectivity index is then assigned to all buildings within each respective plot, reflecting the accessibility of essential services. This approach effectively incorporates facility importance and spatial distribution while utilizing a geographically weighted method to quantify site accessibility.

Additionally, a functional diversity metric is introduced to evaluate the heterogeneity and distribution balance of building functions within a site. First, building classification and urban planning datasets are processed to ensure geometric validity, and building centroids are computed to facilitate spatial association. Through spatial connectivity operations, building function data is linked to plot planning information, enabling the identification of functional attributes at both building and site levels.

\subsubsection{Public Transit Accessibility}

To evaluate transit accessibility, we developed a method integrating road networks, land parcel centroids, and public transport facilities. Spatial data layers—including roads, bus stops, MRT stations, and plots—were projected to a common coordinate system (EPSG:3414), and the road network was transformed into a graph for network-based walking distance analysis.

Each plot centroid serves as the origin for calculating walking access to nearby facilities. We distinguish between \textbf{bus stops} and \textbf{MRT stations} using mode-specific thresholds: 250 meters for bus stops and 500 meters for MRT stations. Transit facilities are considered accessible if their walking distance from the centroid falls within the respective threshold.

The public transit accessibility score $A$ is defined as:

\[
A = \frac{\text{Number of Accessible Facilities}}{\text{Total Facilities Within Thresholds}}
\]

where “accessible facilities” are bus stops within 250\,m and MRT stations within 500\,m walking distance. The denominator includes all facilities within the corresponding straight-line buffers (250\,m for bus stops, 500\,m for MRT).

Accessibility scores are exported in GeoJSON format for spatial integration. This mode-specific approach offers a more realistic assessment of transit coverage and supports data-informed design decisions.

\subsection{Land Use Intensity}
\label{subsec6}

Land use intensity is a fundamental aspect of urban planning, traditionally assessed through well-established metrics that provide an objective and globally recognized evaluation of development density. Although these indicators have been widely used in conventional site planning frameworks, their universal applicability across various urban contexts justifies their inclusion in the proposed Site Planning Layout Indicator (SPLI) system. This study adopts two key metrics to quantify land development intensity: Floor Area Ratio (FAR) and Building Coverage Ratio (BCR), both derived from architectural and urban planning principles to systematically measure spatial utilization.

Floor Area Ratio (FAR) is computed as the ratio of buildings' total floor area to the plot area:
\[
\text{FAR} = \frac{\text{Total Building Floor Area}}{\text{Plot Area}}
\]
This metric provides a standardized measure of building density and is sourced from architectural design documents or GIS databases.

Building Coverage Ratio (BCR) represents the proportion of land occupied by building footprints relative to the total plot area:
\[
\text{BCR} = \frac{\text{Building Footprint Area}}{\text{Plot Area}} \times 100\%
\]
This indicator, extracted from building morphology datasets or satellite imagery, effectively quantifies ground-level site utilization.

To ensure data consistency and usability in subsequent modeling and analysis, these indicators undergo spatial analysis and preprocessing using GIS software. The standardization and integration of datasets are achieved through GeoPandas, facilitating compatibility across different data sources and enhancing analytical precision within the SPLI framework.

\section{Examples of Data and Analyses}
\label{sec5}

\subsection{Indicator Analysis by Land Use Type}
\label{subsec1}

Different land use types exhibit significant variations in terms of building morphology, layout patterns, functional diversity, public transit accessibility, and land use intensity. A systematic quantitative analysis of these spatial indicators not only reveals the organizational characteristics of urban land parcels but also validates the applicability and effectiveness of the Site Planning Layout Indicator (SPLI) in classifying and interpreting built environment typologies. It is important to note that, due to the intrinsic nature of \textit{Road} parcels, which do not involve the construction of buildings, this category is excluded from the present analysis. Against this background, this section systematically examines five key indicators to evaluate the capacity of SPLI in representing multidimensional urban functional characteristics.

\subsubsection{Building Form}

\begin{figure}[t]
    \centering
    \includegraphics[width=1\linewidth]{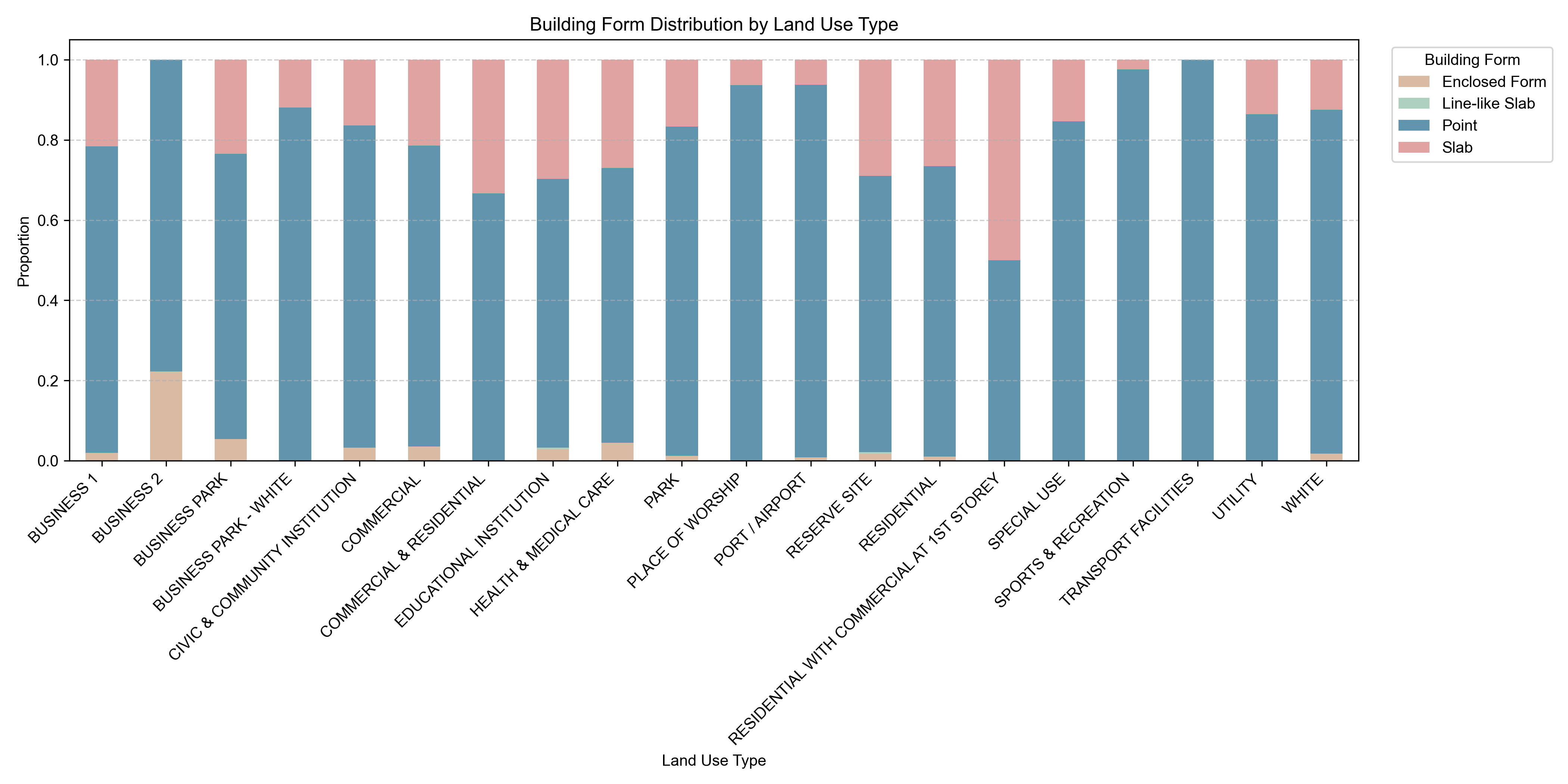}
    \caption{Analysis of building form distribution by land use type}
    \label{fig:AnalysisofBuildingFormDistributionbyLandUseType}
\end{figure}

As shown in Fig.~\ref{fig:AnalysisofBuildingFormDistributionbyLandUseType}, different land use types demonstrate clear distinctions in building form distribution, suggesting that SPLI can effectively capture spatial organizational characteristics. Point-type buildings dominate across most land use types, especially in BUSINESS 1, BUSINESS 2, BUSINESS PARK, PLACE OF WORSHIP, PORT / AIRPORT, TRANSPORT FACILITIES, and WHITE parcels, where their proportion exceeds 80\%. This indicates the high adaptability of point forms in high-rise development or functionally dispersed layouts. In contrast, slab-type buildings are predominantly found in mixed-use parcels such as RESIDENTIAL WITH COMMERCIAL AT 1ST STOREY and COMMERCIAL \& RESIDENTIAL, aligning with the need for linear organization and multi-functional integration. Although enclosed forms constitute a relatively small proportion overall, they are concentrated in EDUCATIONAL INSTITUTION, HEALTH \& MEDICAL CARE, BUSINESS 2, and BUSINESS PARK parcels, suggesting their suitability for clearly organized, pedestrian-accessible, and spatially integrated environments. Overall, the close correlation between building form and land use type underscores the classification capability of SPLI in recognizing spatial morphological structures.

\subsubsection{Layout Patterns}

\begin{figure}[t]
    \centering
    \includegraphics[width=1\linewidth]{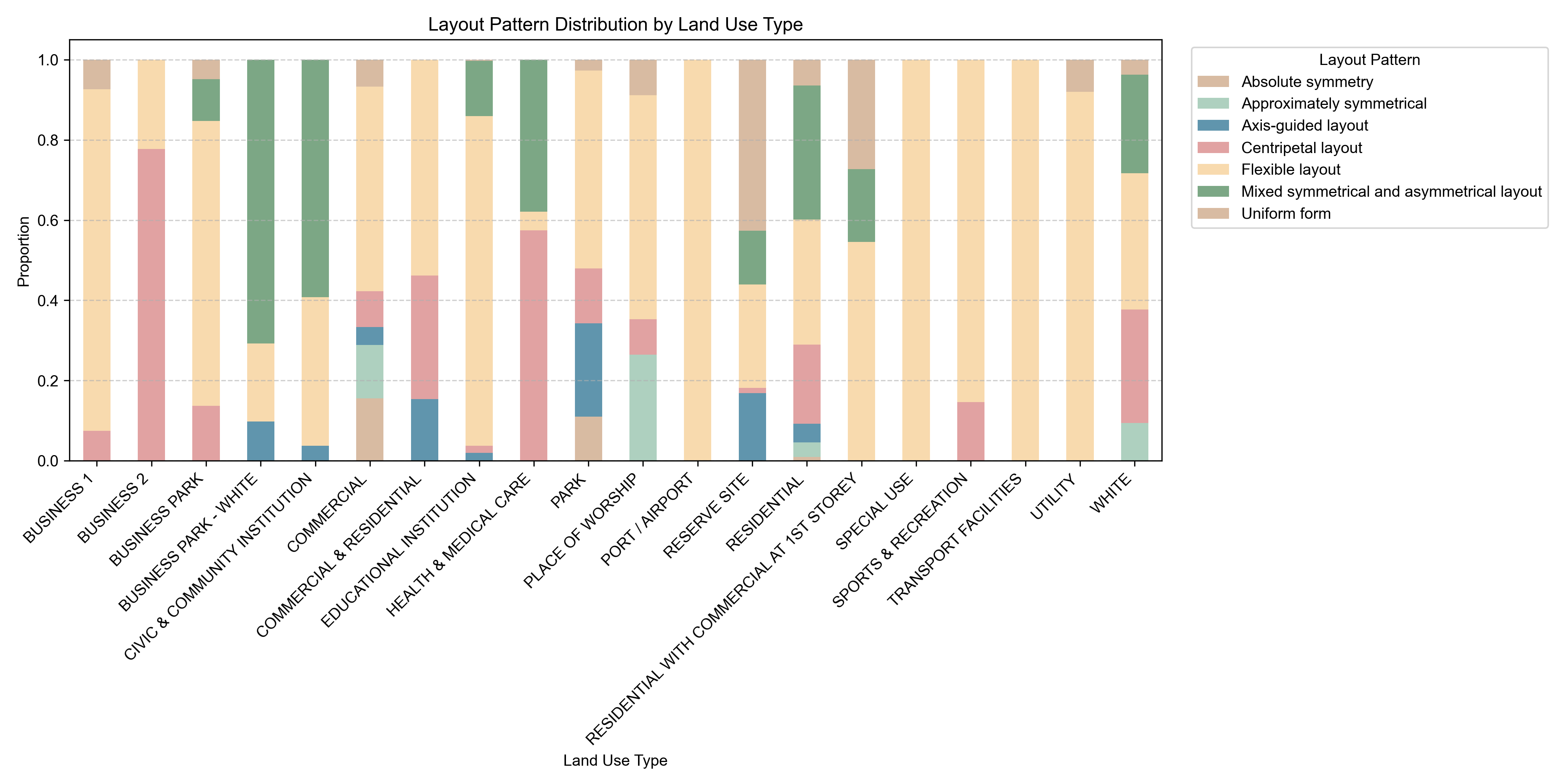}
    \caption{Analysis of layout pattern distribution by land use type}
    \label{fig:AnalysisofLayoutPatternDistributionbyLandUseType}
\end{figure}

As shown in Fig.~\ref{fig:AnalysisofLayoutPatternDistributionbyLandUseType}, the distribution of building layout patterns varies considerably across land use types, highlighting the intrinsic relationship between spatial organization strategies and functional attributes. Flexible Layout is the most prevalent pattern and is widely observed in BUSINESS 1, BUSINESS PARK, EDUCATIONAL INSTITUTION, PORT / AIRPORT, SPECIAL USE, SPORTS \& RECREATION, TRANSPORT FACILITIES, and UTILITY parcels. This reflects a development tendency toward adaptability and non-linear spatial arrangements in these areas. Mixed Symmetrical and Asymmetrical Layout is common in CIVIC \& COMMUNITY INSTITUTION, EDUCATIONAL INSTITUTION, and HEALTH \& MEDICAL CARE parcels, which typically balance order with functional versatility. Axis-Guided Layout appears in PLACE OF WORSHIP, COMMERCIAL \& RESIDENTIAL, and PARK parcels, indicating a reliance on spatial axes for functional organization. Absolute Symmetry and Approximate Symmetry layouts are more prominent in COMMERCIAL, PLACE OF WORSHIP, RESERVE SITE, and RESIDENTIAL parcels, where spatial order and formal presence are emphasized to create a sense of hierarchy and identity. Centripetal Layout, often used to enhance internal cohesion and spatial focus, is mostly found in HEALTH \& MEDICAL CARE, BUSINESS 2, and RESIDENTIAL parcels. These findings further reinforce SPLI’s utility in capturing the structural characteristics of functional spatial arrangements.

\subsubsection{Functional Diversity}

\begin{figure}[t]
    \centering
    \includegraphics[width=1\linewidth]{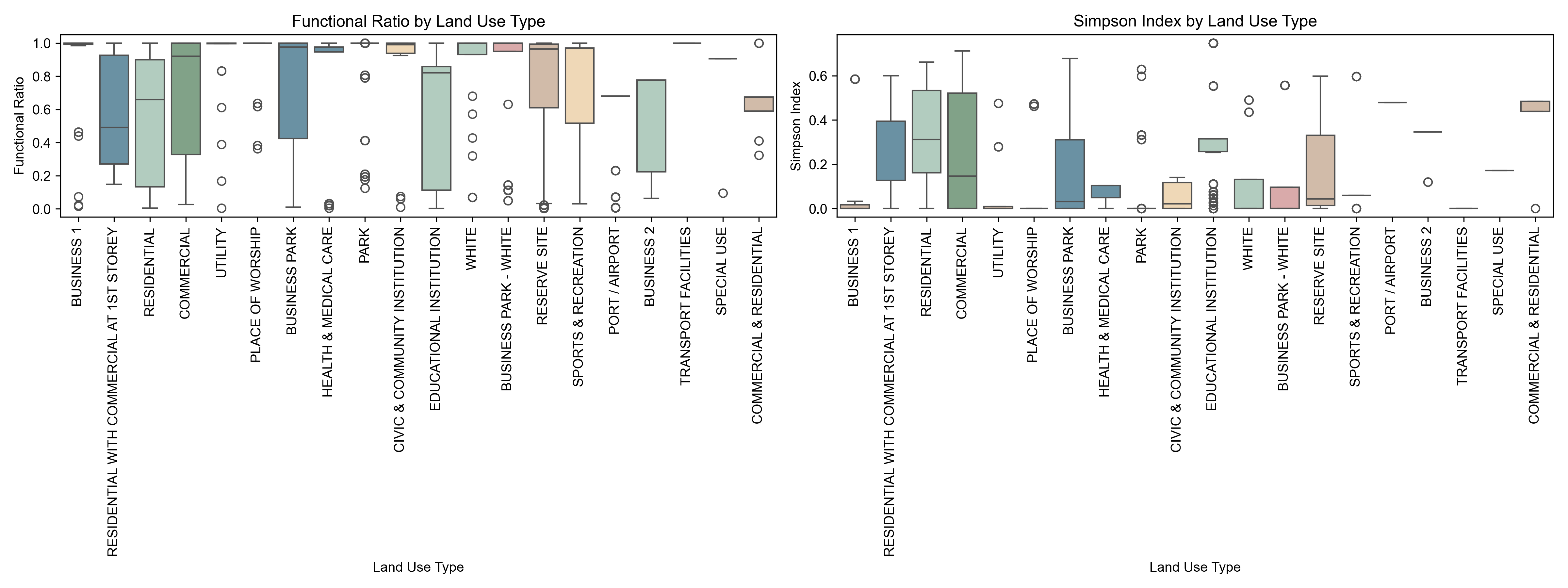}
    \caption{Analysis of functional diversity by land use type}
    \label{fig:AnalysisofFunctionalDiversitybyLandUseType}
\end{figure}

As illustrated in Fig.~\ref{fig:AnalysisofFunctionalDiversitybyLandUseType}, most land parcels exhibit a highly concentrated distribution in terms of functional ratio. In BUSINESS 1, UTILITY, SPORTS \& RECREATION, and TRANSPORT FACILITIES parcels, the median functional ratio is close to or equal to 1, reflecting a high degree of single-function dominance. In contrast, EDUCATIONAL INSTITUTION, HEALTH \& MEDICAL CARE, and RESIDENTIAL parcels display greater dispersion, indicating higher degrees of functional mix and spatial diversity. Regarding the Simpson Index, the median values are highest in COMMERCIAL, RESIDENTIAL, and COMMERCIAL \& RESIDENTIAL parcels (ranging from 0.4 to 0.6), suggesting stronger functional diversity. On the other hand, BUSINESS 1, PLACE OF WORSHIP, and HEALTH \& MEDICAL CARE parcels generally exhibit Simpson Index values below 0.1, indicating a more singular functional composition. These results validate the sensitivity and practicality of SPLI in differentiating between mono-functional and multi-functional spaces.

\subsubsection{Essential Services Accessibility}

\begin{figure}[t]
    \centering
    \includegraphics[width=1\linewidth]{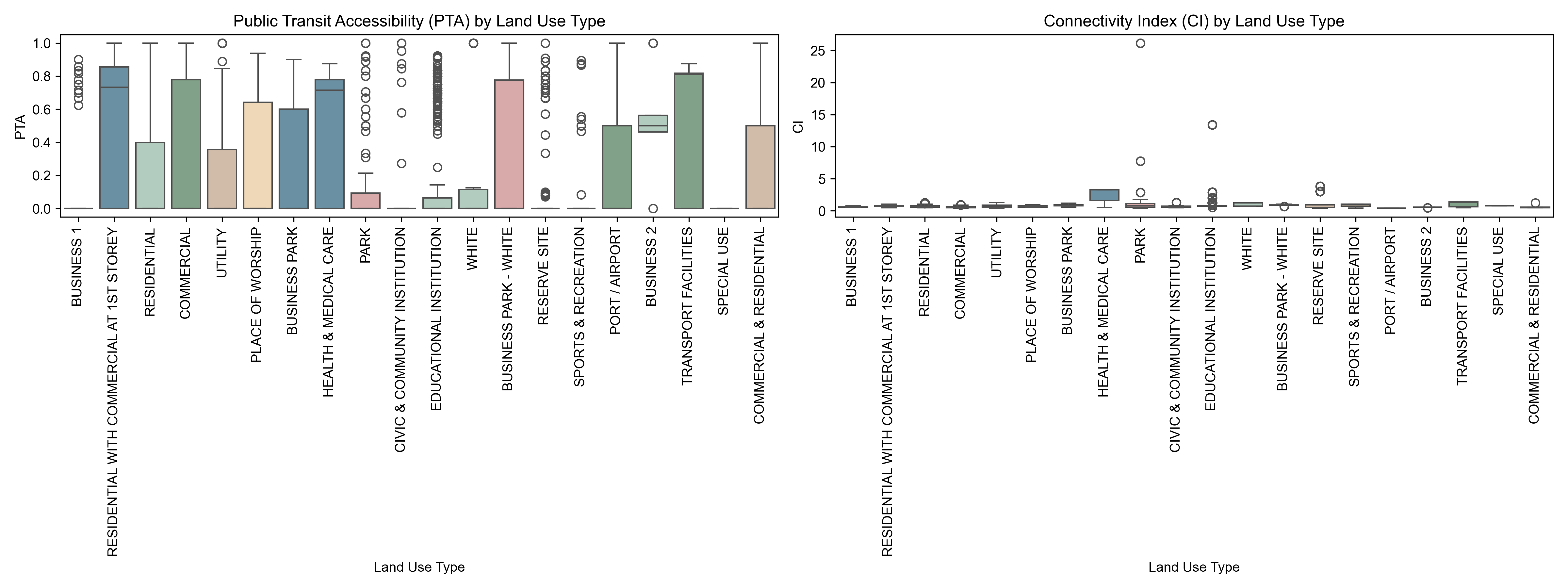}
    \caption{Analysis of public transit accessibility and connectivity by land use type}
    \label{fig:AnalysisofPublicTransitAccessibilityConnectivitybyLandUseType}
\end{figure}

Public transit accessibility metrics, as visualized in Fig.~\ref{fig:AnalysisofPublicTransitAccessibilityConnectivitybyLandUseType}, show notable differences across land use types. RESIDENTIAL WITH COMMERCIAL AT 1ST STOREY, COMMERCIAL, BUSINESS PARK – WHITE, and TRANSPORT FACILITIES parcels have PTA median values ranging between 0.75 and 0.85, indicating good public transit coverage and accessibility. In contrast, parcels such as PARK, EDUCATIONAL INSTITUTION, and UTILITY show median PTA values below 0.3, revealing relatively limited access to public transit. Regarding the Connectivity Index (CI), most parcels have median values between 0.5 and 2.0. However, some outliers—such as HEALTH \& MEDICAL CARE, BUSINESS PARK, and PARK—display unusually high maximum values (25.7, 9.8, and 6.9, respectively), which likely correspond to localized high-connectivity nodes or dense network areas. These extremes further highlight SPLI’s ability to detect and characterize complex spatial network structures.

\subsubsection{Land Use Intensity}

\begin{figure}[t]
    \centering
    \includegraphics[width=1\linewidth]{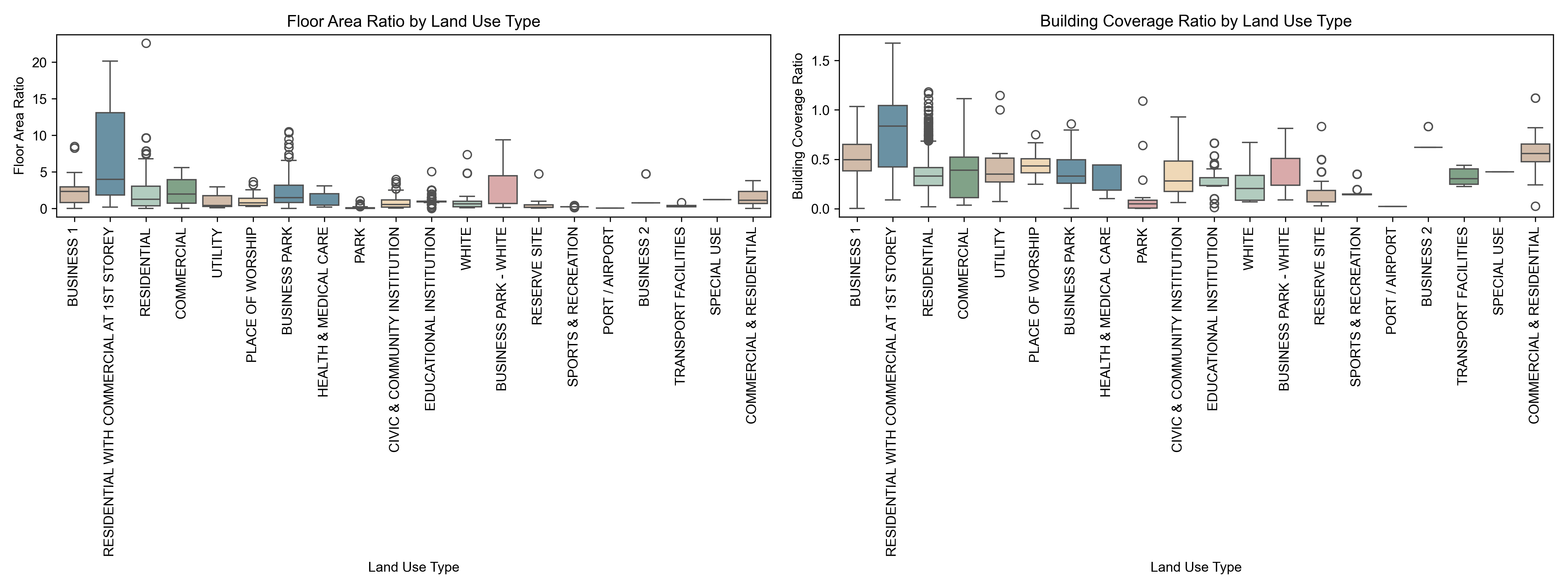}
    \caption{Analysis of land use intensity  by land use type}
    \label{fig:AnalysisofLanduseintensitybyLandUseType}
\end{figure}

Fig.~\ref{fig:AnalysisofLanduseintensitybyLandUseType} presents the distribution of land use intensity across different parcel types. In terms of Floor Area Ratio (FAR), the median value for RESIDENTIAL WITH COMMERCIAL AT 1ST STOREY parcels is significantly higher (approximately 8.0), indicating a high-intensity mixed-use development model. In comparison, RESIDENTIAL and COMMERCIAL parcels show median FAR values around 2.5 and 3.0, respectively, corresponding to medium-density development. Functional parcels such as EDUCATIONAL INSTITUTION, WHITE, and BUSINESS PARK typically have FAR values below 2.0, while PARK, TRANSPORT FACILITIES, and PORT / AIRPORT parcels approach 0, reflecting their open and low-density nature. Regarding Building Coverage Ratio (BCR), RESIDENTIAL and RESIDENTIAL WITH COMMERCIAL AT 1ST STOREY parcels display high median values (around 0.7). However, some samples exhibit BCR values exceeding 1.0, which are treated as outliers caused by data inconsistencies—such as the inclusion of protruding structures (e.g., porches and covered walkways) as part of the building mass in the Master Plan 2019 and OSM data. These are excluded from median calculations. Public-use parcels such as CIVIC \& COMMUNITY INSTITUTION and PLACE OF WORSHIP typically show BCR medians between 0.4 and 0.6, whereas PARK and RESERVE SITE parcels tend to have BCR values below 0.1. In summary, both FAR and BCR effectively capture the development intensity of various land use types, affirming SPLI’s classification sensitivity and representational capability within this dimension.

\subsection{Analysis at the Subzone Level}
\label{subsec2}

At the urban scale, the Site Planning Layout Indicator (SPLI) provides a systematic and quantitative approach for understanding the spatial configuration and functional characteristics within different subzones. In contrast to conventional urban analysis methods that often focus on isolated indicators, SPLI integrates key spatial variables—including building form, layout pattern, functional diversity, accessibility, and land use intensity—to construct a multidimensional representation framework of the built environment. Based on empirical data from multiple subzones in Singapore, this study systematically evaluates the applicability of SPLI in quantifying urban functional layout and further explores its potential in supporting spatial strategy formulation and urban planning decision-making.

\subsubsection{Building Form}
 
\begin{figure}[t]
    \centering
    \includegraphics[width=1\linewidth]{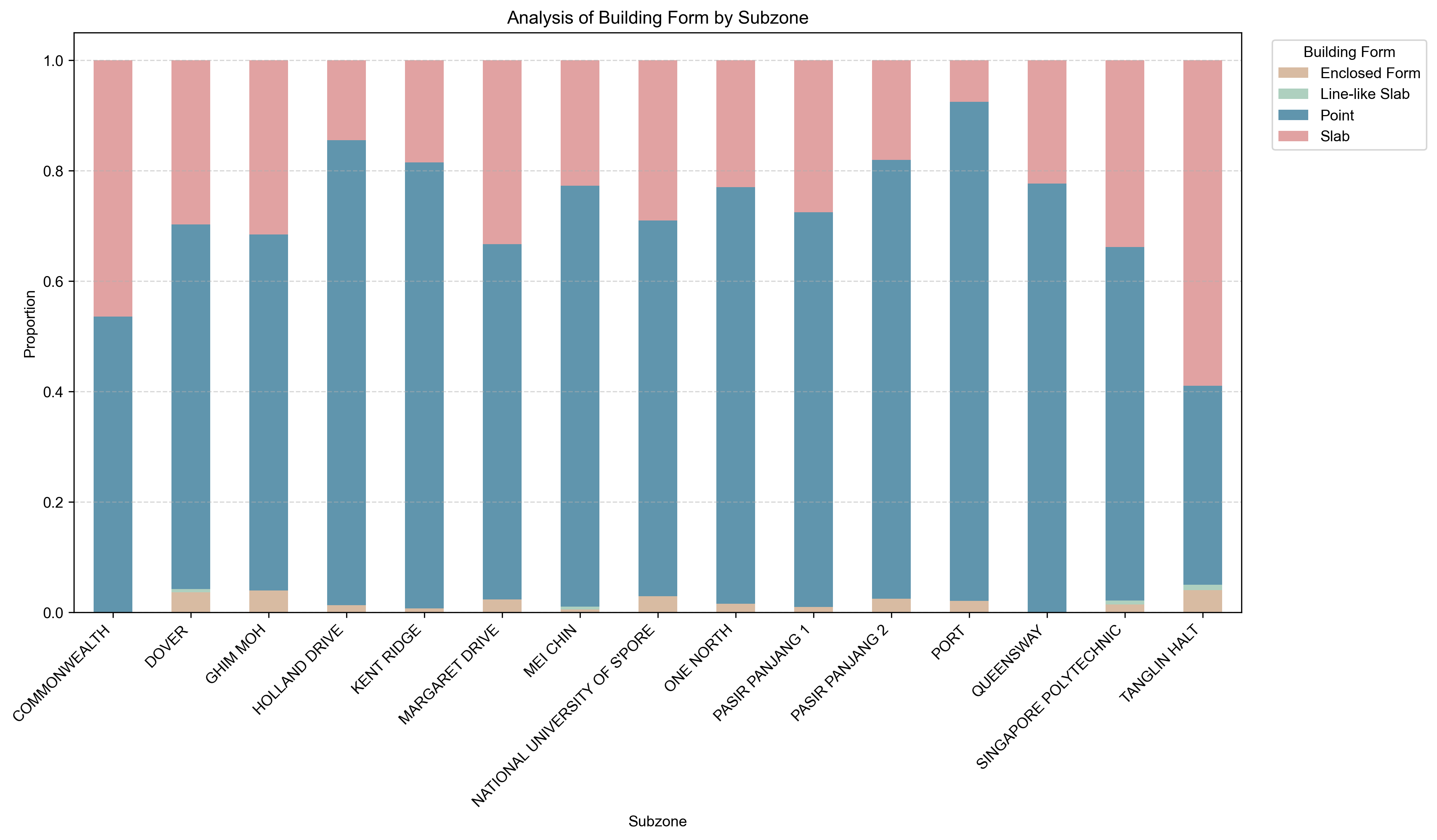}
    \caption{Analysis of building form by subzone}
    \label{fig:AnalysisofBuildingFormbySubzone}
\end{figure}

As shown in Fig.~\ref{fig:AnalysisofBuildingFormbySubzone}, point-type buildings are the most dominant morphological category across most subzones. This is particularly evident in Kent Ridge, Pasir Panjang 1 and 2, Ghim Moh, and Holland Drive, where such forms constitute a significantly higher proportion compared to other types. These areas are generally characterized by high-rise towers or freestanding buildings, indicating a pronounced tendency toward vertical development. The Port area also displays an exceptionally high proportion of point-type buildings, likely reflecting its functionally driven emphasis on storage, logistics, and spatial flexibility. In contrast, slab-type buildings are mainly found in Commonwealth, Tanglin Halt, and Dover, with Tanglin Halt being especially notable—slab buildings account for over 50\% of the stock, reflecting the area's standardized HDB layout, which prioritizes linear arrangement and ventilation efficiency. Enclosed forms are concentrated in education and research precincts such as Singapore Polytechnic, the National University of Singapore (NUS), and One North, suggesting the suitability of inward-facing spatial configurations for ensuring internal cohesion and walkability in institutional environments.

\subsubsection{Layout Patterns}  
\begin{figure}[t]
    \centering
    \includegraphics[width=1\linewidth]{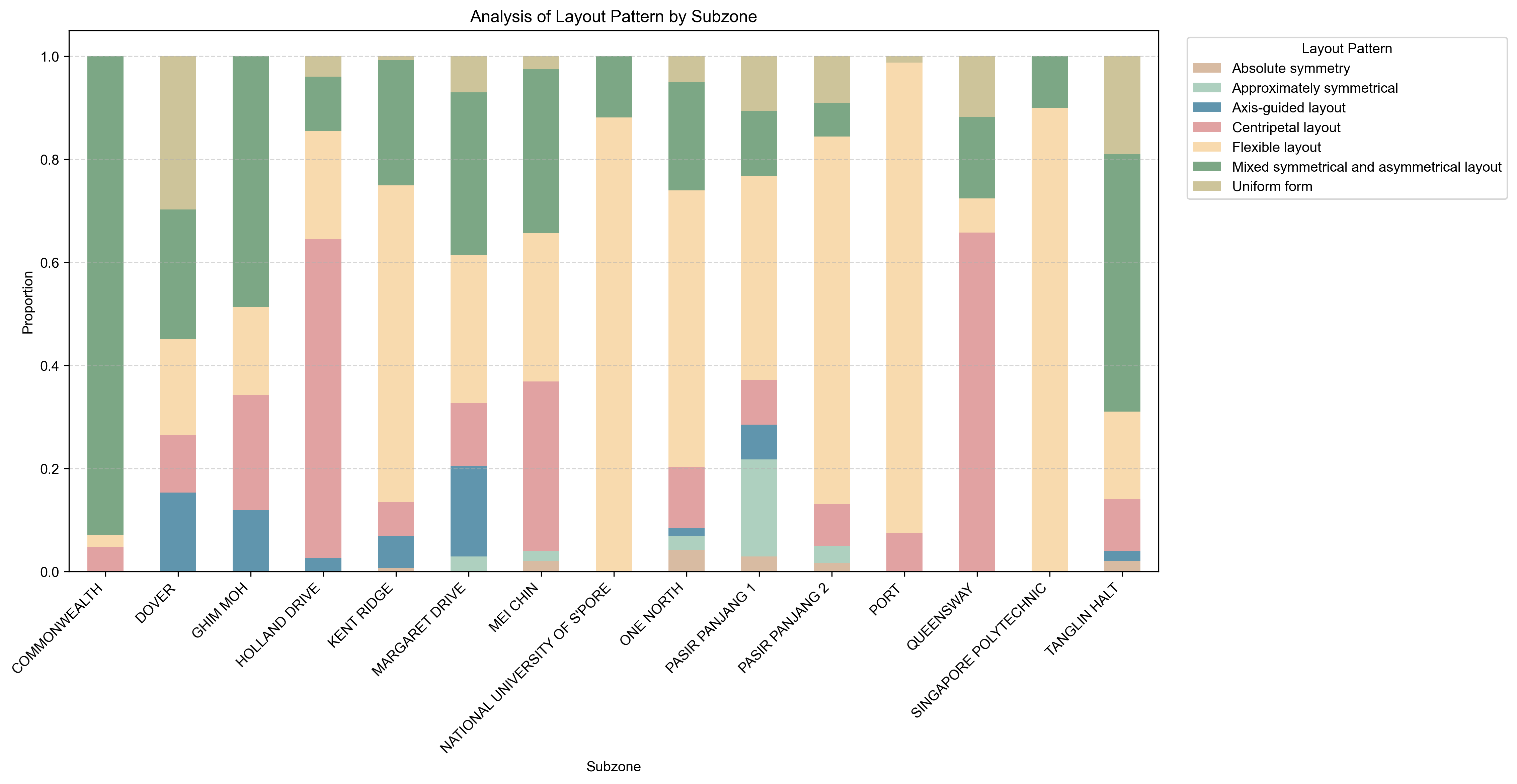}
    \caption{Analysis of layout patterns by subzone}
    \label{fig:AnalysisofLayoutPatternbySubzone}
\end{figure}

As illustrated in Fig.~\ref{fig:AnalysisofLayoutPatternbySubzone}, flexible layouts dominate the spatial configuration of most subzones, especially in One North, NUS, Singapore Polytechnic, and Pasir Panjang 2, where non-linear, adaptable structures prevail. In contrast, Dover and Tanglin Halt exhibit higher proportions of absolutely or approximately symmetrical patterns, reflecting a more regular and orderly planning logic. Axis-guided layouts are relatively prominent in Kent Ridge and One North, indicating spatial organization structured along functional or visual axes. Commonwealth is primarily characterized by a mix of symmetrical and asymmetrical layouts, representing an evolution from standardized planning to more locally responsive configurations. Overall, SPLI proves capable of capturing spatial organization patterns across subzones, highlighting its robust classification and typology identification ability at the urban scale.

\subsubsection{Functional Diversity}  
\begin{figure}[t]
    \centering
    \includegraphics[width=1\linewidth]{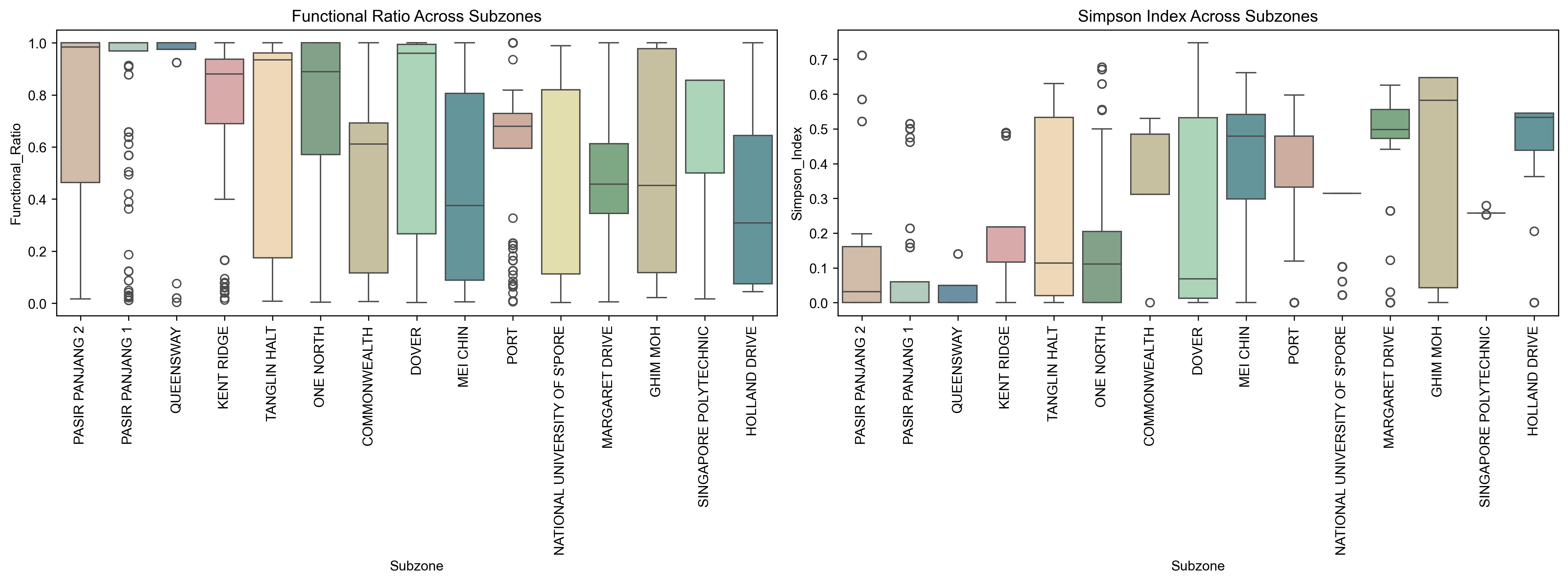}
    \caption{Analysis of functional diversity by subzone}
    \label{fig:AnalysisofFunctionalDiversitybySubzone}
\end{figure}

As shown in Fig.~\ref{fig:AnalysisofFunctionalDiversitybySubzone}, the functional ratio in Pasir Panjang 1, Queensway, and Kent Ridge is highly concentrated, with narrow boxplots and median values approaching 1.0. This indicates a strong dominance of single-function building clusters—typically associated with institutional, industrial, or transport-related land uses. Conversely, Tanglin Halt, Commonwealth, and Dover show more dispersed distributions, including several low values and outliers, suggesting more heterogeneous functional compositions and a degree of mixed-use development. In terms of the Simpson Index, subzones such as Holland Drive, Ghim Moh, and Dover exhibit significantly higher medians and upper quartiles, indicating a more balanced distribution of function types and greater spatial diversity. By contrast, Pasir Panjang 1/2, Queensway, and Kent Ridge all show Simpson Index values below 0.1, pointing to highly concentrated and functionally homogeneous spatial patterns. These results confirm the sensitivity of SPLI in distinguishing between mono-functional and mixed-use configurations.

\subsubsection{Essential Services Accessibility}  
\begin{figure}[t]
    \centering
    \includegraphics[width=1\linewidth]{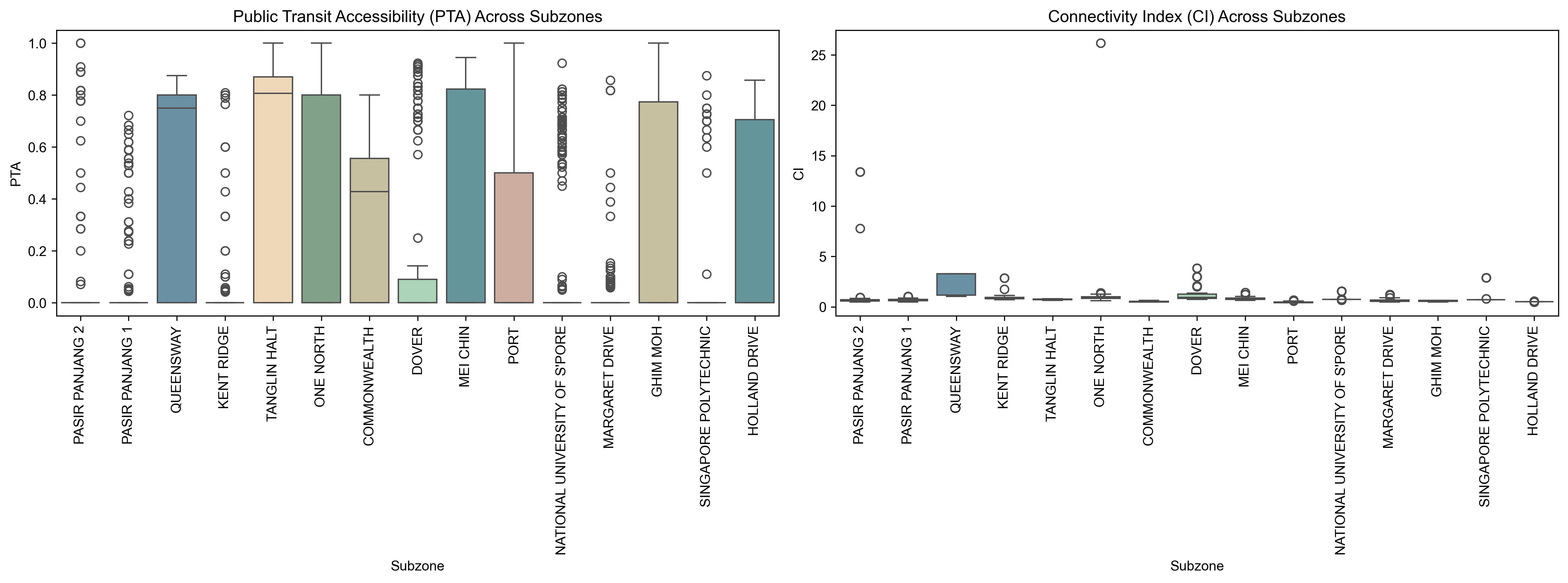}
    \caption{Analysis of public transit accessibility and connectivity by subzone}
    \label{fig:AnalysisofAccessibilitytoEssentialServicesbySubzone}
\end{figure}

As depicted in Fig.~\ref{fig:AnalysisofAccessibilitytoEssentialServicesbySubzone}, the Public Transit Accessibility (PTA) index ranges from 0.6 to 0.9 in most subzones, with Tanglin Halt, Ghim Moh, Mei Chin, and Queensway all having median values above 0.8, indicating well-connected public transport networks. In contrast, PTA values in Pasir Panjang 1/2 and Singapore Polytechnic are significantly lower, with some samples approaching zero—revealing a lack of effective transit nodes within walkable distance. The Connectivity Index (CI) values mostly fall between 0.1 and 1.5, reflecting limited accessibility to diverse amenities within most subzones. However, subzones such as Queensway and Dover exhibit multiple extreme values (exceeding 3 and 5, respectively, and up to 26 in some parcels), likely associated with clusters near commercial hubs or major transit interchanges. These outliers illustrate the capability of SPLI to detect highly aggregated and functionally intensive urban nodes.

\subsubsection{Land Use Intensity}

\begin{figure}[t]
    \centering
    \includegraphics[width=1\linewidth]{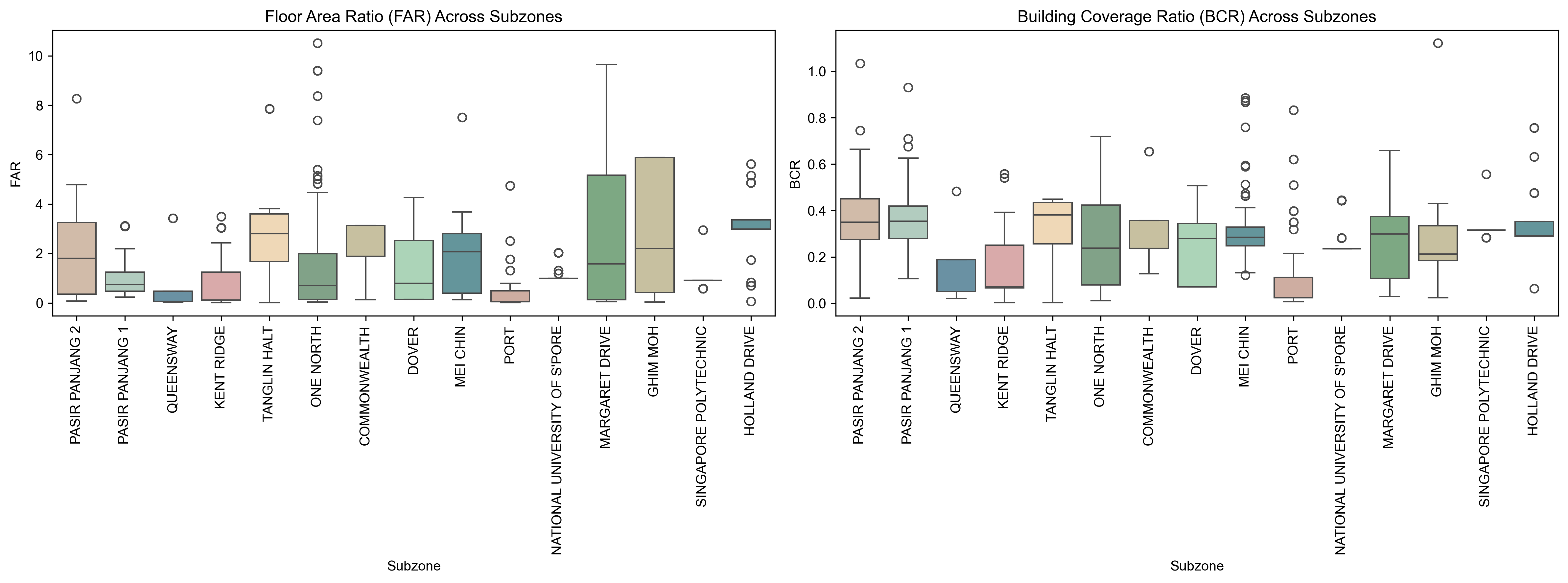}
    \caption{Analysis of land use intensity by subzone}
    \label{fig:AnalysisofLandUseIntensitybySubzone}
\end{figure}

Fig.~\ref{fig:AnalysisofLandUseIntensitybySubzone} presents the distribution of Floor Area Ratio (FAR) and Building Coverage Ratio (BCR) as two key indicators of land use intensity across subzones. Holland Drive shows a markedly higher FAR median of approximately 4.0, with a maximum close to 6.0, suggesting high-density and concentrated development. Ghim Moh and Tanglin Halt follow, both with FAR medians above 3.0. In contrast, subzones such as Queensway, the Port area, and Singapore Polytechnic exhibit FAR medians below 1.0, indicative of low-density development. Regarding BCR, most subzones display medians in the 0.2–0.4 range, though some, such as Ghim Moh, contain extreme values exceeding 1.0. These anomalies are primarily due to data artifacts from Master Plan 2019 and OSM, where protruding architectural elements—such as porches or covered walkways—were counted as part of the main building footprint, resulting in inflated BCR values. These outliers are therefore excluded from statistical analysis. In summary, the combined distribution of FAR and BCR effectively captures the real differences in development intensity across subzones, and further confirms SPLI’s high sensitivity in responding to volumetric and density patterns, even in the presence of morphological data inconsistencies.

\section{Discussion}
\label{sec6}

\subsection{Data Availability and Challenges in Representing the Built Environment}

In urban planning and architectural design, high-quality data-driven methodologies have become essential for optimizing spatial configurations. However, existing data structures still present significant limitations, restricting the systematic representation of the built environment. First, while primary data sources such as OpenStreetMap (OSM), Points of Interest (POI), building morphology, land use data, and satellite imagery provide fundamental information on urban functions, they are often characterized by coarse granularity, high heterogeneity, and inconsistent update frequencies, making it challenging to capture the dynamic evolution of urban spaces comprehensively. Additionally, while Geographic Information System (GIS) data is predominantly stored in structured formats, unstructured data sources such as social media records and mobility trajectory data offer unique insights into urban spatial behaviors. However, the integration mechanisms between these datasets remain underdeveloped, limiting the applicability of data-driven analytical methods.

The scale of data selection is equally critical in site planning research. At the micro-scale (e.g., individual buildings), data can optimize architectural layout and functional allocation. At the meso-scale (e.g., neighborhood level), data facilitates analyses of street vitality and public space utilization. At the macro-scale (e.g., city-wide planning), data supports the evaluation of urban development patterns. However, establishing a coherent analytical framework across these scales remains a major challenge, as data analysis must simultaneously inform localized optimization while supporting broader urban planning strategies.

Moreover, data representativeness is a key concern in urban planning research. While data is often perceived as an objective reflection of spatial usage, inherent biases in data collection can significantly influence planning decisions. For instance, POI data is predominantly static, whereas urban activities exhibit high levels of temporal variability. Relying solely on static datasets for functional classification may fail to accurately reflect the actual usage of urban spaces. Therefore, in site planning research, beyond leveraging existing data sources, advanced methodologies such as Graph Neural Networks (GNN) modeling and spatiotemporal analysis are essential to enhance data applicability and predictive capabilities. This ensures that data-driven planning approaches can effectively capture the dynamic evolution of urban environments.

\subsection{Limitations of Existing Methods: The Gap Between Data and Spatial Logic}

Current urban planning approaches predominantly rely on static, land-use-based functional classifications, wherein urban spaces are categorized into residential, commercial, or industrial zones based on administrative boundaries or land-use types. However, this single-dimensional classification fails to accurately represent the actual spatial utilization patterns. For instance, a commercial plot may serve vastly different functions—such as office space, retail, or social activities—at different times of the day. Existing planning frameworks lack the flexibility to dynamically adjust land-use designations, leading to inefficient spatial resource utilization. Furthermore, urban spatial functions are inherently interactive, yet conventional planning methods primarily focus on individual plots or buildings, neglecting the interdependencies between land parcels and their impact on urban operations. For example, the spatial relationship between residential and commercial areas directly affects accessibility and quality of life, while the integration of industrial parks with surrounding service facilities determines economic vibrancy.

Graph Neural Networks (GNNs), particularly Relational Graph Convolutional Networks (RGCNs), offer a novel computational approach to address these challenges. By learning the complex relationships between buildings, roads, POIs, and land use, RGCNs enhance the accuracy of building function prediction and compensate for data gaps. Even when functional data for certain parcels is missing, RGCNs can infer latent land-use patterns by leveraging relational attributes of adjacent buildings. Moreover, RGCNs facilitate multi-scale data integration within a unified analytical framework, improving the precision of spatial reasoning. Building on this foundation, future urban planning models can further integrate spatiotemporal forecasting techniques to enable dynamic land-use adjustments, enhancing the adaptability and responsiveness of planning strategies.

\subsection{Next-Generation Design: Data-Driven Optimization and Intelligent Planning}

With the evolution of data-driven methodologies, urban planning is advancing toward greater intelligence and adaptability. The Site Planning Layout Indicator (SPLI) framework proposed in this study establishes a standardized system of indicators, enabling automated spatial reasoning. However, the dynamic nature of urban development underscores the limitations of static planning approaches. Future research should focus on enhancing the dynamic adaptability of the SPLI framework and expanding its applications in intelligent urban planning.

Integrating SPLI with a knowledge graph representing urban morphological data and functional attributes can significantly enhance its capabilities in spatial reasoning and geospatial querying. Such a graph provides a structured representation of urban relationships, facilitating automated planning decisions based on natural language queries. For instance, a query such as ``Identify residential plots with a Floor Area Ratio (FAR) > 2.5 near commercial centers'' enables the system to rapidly retrieve qualifying parcels and optimize layouts based on accessibility, functional diversity, and surrounding land use.

Incorporating spatiotemporal data analysis further allows for the development of adaptive planning mechanisms, enabling dynamic land-use adjustments in response to evolving demands. For example, during periods of increased commercial activity, certain office spaces can be temporarily repurposed for retail or service functions to maximize spatial efficiency. Similarly, public spaces may be adapted for temporary markets or cultural events during holidays to enhance urban vitality. Through spatiotemporal graph neural networks (ST-GCNs), SPLI can analyze temporal variations in land use patterns, optimize functional layouts, and improve urban resource allocation efficiency.

Furthermore, integrating Large Language Models (LLMs) with Retrieval-Augmented Generation (RAG) can further enhance the intelligence of the SPLI framework in urban planning. Current LLMs primarily rely on textual data and lack deep spatial reasoning capabilities. However, by incorporating SPLI data within RAG-enhanced models, their ability to retrieve and infer geospatial information can be significantly improved. For example, when prompted with "Which plots are suitable for a new school?", the system can analyze SPLI attributes and generate recommendations based on accessibility, land-use classification, and population density. By leveraging LLMs and RAG, the SPLI framework can support standardized geospatial queries while enabling end-to-end intelligent workflows, transforming data-driven insights into actionable planning and design optimization strategies. This integration paves the way for automated architectural design and intelligent urban planning, ensuring more precise, adaptable, and data-driven decision-making processes.

\subsection{Data Integration: Interdisciplinary Applications and Future Potential}

The standardized framework of the Site Planning Layout Indicator (SPLI) not only enhances the structured storage of site planning data but also provides a robust foundation for broader urban studies. In the future, SPLI can be further integrated with healthcare, socio-economic, and environmental-climatic datasets to expand its applicability and support more precise urban planning and policy-making.

Integrating SPLI with a knowledge graph that encodes urban morphological features and functional attributes can substantially enhance its capacity for spatial reasoning and geospatial query handling. By providing a structured representation of urban relationships, such a system enables automated planning decisions driven by natural language queries—for example, identifying residential plots with a Floor Area Ratio (FAR) above 2.5 near commercial centers. Furthermore, incorporating spatiotemporal data facilitates the development of adaptive planning mechanisms, allowing dynamic land-use adjustments in response to shifting demands. During periods of intensified commercial activity, for instance, office spaces could be temporarily reallocated to retail or service functions to optimize spatial efficiency. Public spaces may similarly be repurposed for markets or cultural events during holidays to enhance urban vitality. Leveraging spatiotemporal graph neural networks (ST-GCNs), the system can analyze temporal variations in land-use patterns, refine functional layouts, and support more responsive and efficient urban resource allocation.

Furthermore, SPLI can contribute to urban environmental and climatic analysis, enhancing strategies for microclimate regulation and urban sustainability. By integrating building morphology data with climatic zoning models, SPLI can refine Local Climate Zone (LCZ) classification, improving urban climate resilience. Additionally, incorporating green space coverage, noise pollution, and other environmental metrics allows for habitability assessments across different spatial configurations, informing urban design optimizations. Looking ahead, SPLI has the potential to evolve into an intelligent urban planning system, leveraging real-time data for adaptive spatial function optimization. By facilitating multi-scale urban transformation analysis, SPLI can provide data-driven insights for advancing sustainable urban development.

\section{Conclusion}
\label{sec7}
This study proposes the Site Planning Layout Indicator (SPLI) framework, which adopts a data-driven approach to establish a standardized representation of urban spatial layouts. By systematically quantifying the spatial characteristics of different land parcels, SPLI enhances the precision and consistency of urban morphological indicators. To address the challenge of urban data gaps, SPLI integrates multimodal data fusion with deep learning techniques, effectively compensating for missing data and improving the completeness and reliability of spatial analysis. The experimental results validate the feasibility of this indicator framework, demonstrating that SPLI accurately captures building morphology, functional layout, accessibility, and land use intensity, thereby providing a standardized analytical toolset for data-driven urban planning. Looking ahead, SPLI can further support the development of the Function-based Knowledge Graph (FKG), serving as a foundational dataset for automated, data-driven urban spatial analysis while expanding its applications in intelligent architectural design and urban planning.

\appendix
\section*{Appendix A: Three-Tier Building Function Classification}

\renewcommand{\arraystretch}{0.5}
\scriptsize
\begin{longtable}{
    >{\centering\arraybackslash}p{4.2cm}
    >{\centering\arraybackslash}p{2.8cm}
    >{\centering\arraybackslash}p{1.0cm}
    >{\centering\arraybackslash}p{2.1cm}
    >{\centering\arraybackslash}p{1.5cm}
}

\caption{Summary of the three-tier Building Function  classification} \label{tab:building_summary} \\
\toprule
\textbf{Level 3 Building function} & \textbf{Level 2 Building type} & \textbf{Level 2 code} & \textbf{Level 1 Buildings category} & \textbf{Land use} \\
\midrule
\endfirsthead
\toprule
\textbf{Level 3 Building function} & \textbf{Level 2 Building type} & \textbf{Level 2 number} & \textbf{Level 1 Buildings category} & \textbf{Land use} \\
\midrule
\endhead
Residential-Housing Units-HDB Properties-annex-corridor & Residential-Housing Units-HDB Properties-annex & B3' & Residential buildings & Residential \\
\cmidrule(lr){1-1} 
Annex &  &  &  &  \\
\cmidrule(lr){1-4} 
Flats & Residential-Housing Units-HDB Properties & B3 &  &  \\
\cmidrule(lr){1-3} 
Annex & Residential-Housing Units-Condominiums and Other Apartments-annex & B2' &  &  \\
\cmidrule(lr){1-1} 
Residential-Housing Units-Condominiums and Other Apartments-annex-corridor &  &  &  &  \\
\cmidrule(lr){1-4} 
Condominium & Residential-Housing Units-Condominiums and Other Apartments & B2 &  &  \\
\midrule 
Townhouse & Residential-Housing Units-Landed Properties & B1 & Street buildings &  \\
\cmidrule(lr){1-1} 
Terrace House &  &  &  &  \\
\cmidrule(lr){1-1} 
Semi-Detached House &  &  &  &  \\
\cmidrule(lr){1-1} 
Detached House &  &  &  &  \\
\cmidrule(lr){1-1} 
Strata-Landed Housing &  &  &  &  \\
\cmidrule(lr){1-4} 
Retirement Housing & Residential-Housing Units-Condominiums and Other Apartments & B2 & Residential buildings &  \\
\cmidrule(lr){1-1} 
Serviced Apartments &  &  &  &  \\
\cmidrule(lr){1-1} 
Student Hostel &  &  &  &  \\
\midrule 
Flats with commercial uses at 1st storey & HDB with commercial uses at 1st storey & AB4 & Residential buildings & Residential with Commercial at 1st storey \\
\cmidrule(lr){1-4} 
Shophouse & Shophouse & B5 & Street buildings &  \\
\cmidrule(lr){1-4} 
Residential Developments(e.g. Flats) & Condominiums and Other Apartments with commercial uses at 1st storey & AB4' & Residential buildings &  \\
\midrule 
Mixed Commercial \&Residential development(e.g. Shopping/Hotel/Office \& Residential) & Commercial \& Residential & AB & Public buildings & Commercial \& Residential \\
\cmidrule(lr){2-3} 
 & Commercial, Office \& Residential & AAB &  &  \\
 \cmidrule(lr){2-3} 
 & Office, Residential \& Educational & ABD &  &  \\
 \cmidrule(lr){2-3} 
 & Commercial, Office , Transport facility \& Residential & AAAB &  &  \\
\midrule 
Offices & Office & A2 & Public buildings & Commercial \\
 \cmidrule(lr){1-1} 
Offices-corridor &  &  &  &  \\
\cmidrule(lr){1-3} 
Mixed Uses (e.g. Office, Shopping, Cinema, Hotel, Flat)  & Commercial \& Office \& Transport facility & AAA &  &  \\
 \cmidrule(lr){2-3} 
 & Commercial \& Office & AA' &  &  \\
  \cmidrule(lr){2-3} 
 & Commercial, Office \& Hotel & AAC &  &  \\
  \cmidrule(lr){2-3} 
 & Commercial \& Transport facility & AA &  &  \\
 \cmidrule(lr){1-3} 
Convention/Exhibition Centre & Commercial & A1 &  &  \\
 \cmidrule(lr){1-1} 
Commercial School &  &  &  &  \\
 \cmidrule(lr){1-1} 
Bank &  &  &  &  \\
 \cmidrule(lr){1-1} 
Market/Food Centre/Restaurant &  &  &  &  \\
 \cmidrule(lr){1-1} 
Cinema &  &  &  &  \\
 \cmidrule(lr){1-1} 
Entertainment &  &  &  &  \\
 \cmidrule(lr){1-1} 
Shopping mall &  &  &  &  \\
 \cmidrule(lr){1-1} 
Foreign Trade Mission/Chancery &  &  &  &  \\
 \cmidrule(lr){1-1} 
Commercial-corridor &  &  &  &  \\
 \cmidrule(lr){1-3} 
Commercial \& Hotel & Commercial \& Hotel & AC &  &  \\
\midrule 
Hotel & Hotel & C & Public buildings & Hotel \\
 \cmidrule(lr){1-1} 
Backpackers’ Hostel &  &  &  &  \\
\midrule 
Business Park & Office & A2 & Public buildings & Business Park \\
 \cmidrule(lr){1-1} 
Science Park &  &  &  &  \\
 \cmidrule(lr){1-1} 
Office-corridor &  &  &  &  \\
\midrule 
Clean and Light industrial factories with one or more predominant uses as listed in the Handbook on Development Control Parameters for industrial  developments. & Commercial \& Industrial & AE' & Public buildings & Business 1 \\
 \cmidrule(lr){2-4} 
 & Industrial & E & Industrial Buildings &  \\
 \cmidrule(lr){2-4} 
 & Office \& Industrial & AE & Public buildings &  \\
\midrule 
Clean, Light and General industrial factories with one or more predominant uses as listed in the Handbook on Development Control Parameters for industrial developments. & Industrial & E & Industrial Buildings & Business 2 \\
\midrule 
Residential Developments (e.g. Flats) & Residential \& Institutional  & AG & Residential buildings & Residential/ Institution \\
 \cmidrule(lr){1-1} 
Community Institution uses (excluding funeral parlour and workers’ dormitory) &  &  &  &  \\
\midrule 
Entertainment & Commercial & A1 & Public buildings & Commercial/Institution \\
 \cmidrule(lr){1-1} 
Recreation Club &  &  &  &  \\
 \cmidrule(lr){1-1} 
Offices &  &  &  &  \\
 \cmidrule(lr){1-1} 
Bank &  &  &  &  \\
 \cmidrule(lr){1-1} 
Shops &  &  &  &  \\
 \cmidrule(lr){1-1} 
Commercial School &  &  &  &  \\
 \cmidrule(lr){1-1} 
Food Centres/Restaurant &  &  &  &  \\
 \cmidrule(lr){1-3} 
Community Institution facilities e.g. child care centres, association premises (excluding funeral parlours and workers’ dormitories) & Institutional  & G &  &  \\
\midrule 
Hospital & Hospital & D & Public buildings & Health \& Medical Care \\
 \cmidrule(lr){1-1} 
Hospital-corridor &  &  &  &  \\
\midrule 
Polyclinic & Clinic & D' &  &  \\
 \cmidrule(lr){1-1} 
Clinic/Dental Clinic &  &  &  &  \\
 \cmidrule(lr){1-1} 
Veterinary Clinic &  &  &  &  \\
 \cmidrule(lr){1-3} 
Nursing Home & Health \& Medical Care & D &  &  \\
 \cmidrule(lr){1-3} 
Medical suite & Hospital & D &  &  \\
\midrule 
Kindergarten / Preschool & Educational & F & Public buildings & Educational Institution \\
 \cmidrule(lr){1-1} 
Primary School &  &  &  &  \\
 \cmidrule(lr){1-1} 
Secondary School &  &  &  &  \\
 \cmidrule(lr){1-1} 
School &  &  &  &  \\
 \cmidrule(lr){1-3} 
Educational-annex & Educational-annex & F' &  &  \\
 \cmidrule(lr){1-1} 
Educational-corridor &  &  &  &  \\
 \cmidrule(lr){1-3} 
Junior College & Educational & F &  &  \\
 \cmidrule(lr){1-1} 
Institute of Technical Education &  &  &  &  \\
 \cmidrule(lr){1-1} 
Polytechnic &  &  &  &  \\
 \cmidrule(lr){1-1} 
University &  &  &  &  \\
 \cmidrule(lr){1-1} 
Religious School/ Institute &  &  &  &  \\
 \cmidrule(lr){1-1} 
Foreign System School &  &  &  &  \\
 \cmidrule(lr){1-1} 
Special Education School (e.g. School for the Disabled) &  &  &  &  \\
\midrule 
Church & Place of Worship & L & Public buildings & Place of Worship \\
\cmidrule(lr){1-1} 
Mosque &  &  &  &  \\
\cmidrule(lr){1-1} 
Temple &  &  &  &  \\
\midrule 
Courts & Institutional & G & Public buildings & Civic \& Community Institution \\
\cmidrule(lr){1-1} 
Police Station &  &  &  &  \\
\cmidrule(lr){1-1} 
Fire Station &  &  &  &  \\
\cmidrule(lr){1-1} 
Prison &  &  &  &  \\
\cmidrule(lr){1-1} 
Drug Rehabilitation Centre/Halfway House &  &  &  &  \\
\cmidrule(lr){1-1} 
Reformative Centre Community Institutions &  &  &  &  \\
\cmidrule(lr){1-1} 
Association premises &  &  &  &  \\
\cmidrule(lr){1-1} 
Community Centre/Club &  &  &  &  \\
\cmidrule(lr){1-1} 
Community Hall &  &  &  &  \\
\cmidrule(lr){1-1} 
Welfare Home &  &  &  &  \\
\cmidrule(lr){1-1} 
Child Care Centre &  &  &  &  \\
\cmidrule(lr){1-1} 
Home For The Aged &  &  &  &  \\
\cmidrule(lr){1-1} 
Home For The Disabled &  &  &  &  \\
\cmidrule(lr){1-1} 
Funeral Parlour/Cemetery &  &  &  &  \\
\cmidrule(lr){1-1} 
Workers’ Dormitory Cultural Institutions &  &  &  &  \\
\cmidrule(lr){1-1} 
Television/Filming Studio Complex &  &  &  &  \\
\cmidrule(lr){1-1} 
Performing Arts Centre &  &  &  &  \\
\cmidrule(lr){1-1} 
Library &  &  &  &  \\
\cmidrule(lr){1-1} 
Museum &  &  &  &  \\
\cmidrule(lr){1-1} 
Arts Centre/Science Centre &  &  &  &  \\
\cmidrule(lr){1-1} 
Concert Hall &  &  &  &  \\
\cmidrule(lr){1-1} 
Institutional-corridor  &  &  &  &  \\
\midrule 
National Park & Park & H & Public buildings & Park \\
\cmidrule(lr){1-1} 
Regional Park &  &  &  &  \\
\cmidrule(lr){1-1} 
Community Park/ Neighbourhood Park &  &  &  &  \\
\cmidrule(lr){1-1} 
Park Connectors &  &  &  &  \\
\cmidrule(lr){1-1} 
Zoological Gardens, Botanic Gardens, etc. &  &  &  &  \\
\midrule 
Sports Complex/ Indoor Stadium & Sports \& Recreation & I & Public buildings & Sports \& Recreation \\
\cmidrule(lr){1-1} 
Swimming Complex &  &  &  &  \\
\cmidrule(lr){1-1} 
Golf Course &  &  &  &  \\
\cmidrule(lr){1-1} 
Golf Driving Range &  &  &  &  \\
\cmidrule(lr){1-1} 
Recreation Club &  &  &  &  \\
\cmidrule(lr){1-1} 
Campsite &  &  &  &  \\
\cmidrule(lr){1-1} 
Chalet &  &  &  &  \\
\cmidrule(lr){1-1} 
Marina &  &  &  &  \\
\cmidrule(lr){1-1} 
Water Sports Centre &  &  &  &  \\
\cmidrule(lr){1-1} 
Outward Bound School &  &  &  &  \\
\cmidrule(lr){1-1} 
Theme Park &  &  &  &  \\
\cmidrule(lr){1-1} 
Sports \& Recreation-corridor &  &  &  &  \\
\midrule 
Car Park & Transport facility & A3 & Public buildings & Transport Facilities \\
\cmidrule(lr){1-1} 
Heavy Vehicle Park &  &  &  &  \\
\cmidrule(lr){1-1} 
Trailer Park &  &  &  &  \\
\cmidrule(lr){1-3} 
Bus Depot/Terminal & Public transportation facility & A5 &  &  \\
\cmidrule(lr){1-3} 
Transport Depot & Transport facility & A3 &  &  \\
\cmidrule(lr){1-3} 
MRT/LRT Marshalling Yard/Depot & Public transportation facility & A5 &  &  \\
\cmidrule(lr){1-3} 
Driving Circuit/Test Centre & Transport facility & A3 &  &  \\
\cmidrule(lr){1-3} 
Bus station & Public transportation facility & A5 &  &  \\
\cmidrule(lr){1-3} 
Petrol Station/Kiosk & Transport facility & A3 &  &  \\
\midrule 
MRT/LRT Station & Public transportation facility & A5 & Public buildings & Rapid transit \\
\midrule 
Electrical Substation & Utility & J & Public buildings & Utility \\
\cmidrule(lr){1-1} 
Power Station &  &  &  &  \\
\cmidrule(lr){1-1} 
Gas Installation &  &  &  &  \\
\cmidrule(lr){1-1} 
Natural Gas Receiving Terminal &  &  &  &  \\
\cmidrule(lr){1-1} 
Gas Takeoff/Regulator Station &  &  &  &  \\
\cmidrule(lr){1-1} 
Water Treatment Plant &  &  &  &  \\
\cmidrule(lr){1-1} 
Water Reclamation Plant &  &  &  &  \\
\cmidrule(lr){1-1} 
Service Reservoir &  &  &  &  \\
\cmidrule(lr){1-1} 
Water Pump House &  &  &  &  \\
\cmidrule(lr){1-1} 
Sewage Pumping Station &  &  &  &  \\
\cmidrule(lr){1-1} 
Incineration Plant &  &  &  &  \\
\cmidrule(lr){1-1} 
Desalination Plant &  &  &  &  \\
\cmidrule(lr){1-1} 
Transmitting Station/ Receiving Station &  &  &  &  \\
\cmidrule(lr){1-1} 
Earth Satellite Station &  &  &  &  \\
\midrule 
Cemetery & Cemetery & O & Public buildings & Cemetery \\
\cmidrule(lr){1-1} 
Crematorium &  &  &  &  \\
\cmidrule(lr){1-1} 
Columbarium &  &  &  &  \\
\midrule 
Airport & Port/Airport & A4 & Tranport buildings & Port / Airport \\
\cmidrule(lr){1-1} 
Port Area &  &  &  &  \\
\cmidrule(lr){1-1} 
Port/Airport Related Facilities &  &  &  &  \\
\cmidrule(lr){1-1} 
Ferry Point/Terminal &  &  &  &  \\
\cmidrule(lr){1-1} 
Cruise Centre &  &  &  &  \\
\cmidrule(lr){1-1} 
Landing Sites &  &  &  &  \\
\cmidrule(lr){1-1} 
Fishing Port &  &  &  &  \\
\cmidrule(lr){1-1} 
Port/Airport-corridor of  &  &  &  &  \\
\midrule 
These are areas used or intended to be used for special purposes. & Nil &  &  & Special Use \\
\cmidrule(lr){1-4} 
Pavilion & Infrastructure & N & Public buildings &  \\
\midrule 
\bottomrule
\end{longtable}
%
%
\bibliographystyle{elsarticle-num}  
\bibliography{references}    

\end{document}